\documentclass{article} 

\usepackage{arxiv}

\usepackage{tikz}
\usepackage{amsmath}

\usepackage{microtype}
\usepackage{graphicx}
\usepackage{caption}
\usepackage{amsfonts}
\usepackage{booktabs} 
\usepackage{subcaption}
\usepackage{siunitx}
\usepackage[hyphenbreaks]{breakurl}
\usepackage{multirow}
\usepackage{subcaption}
\usepackage{graphicx}
\graphicspath{ {pics/} {pics/re152advdist} {pics/vggdist} {pics/throttleeffect}}
\usepackage{xcolor}

\usepackage{color}
\usepackage[numbers,square]{natbib}
\usepackage[colorlinks = true,
            linkcolor = purple,
            urlcolor  = blue,
            citecolor = cyan,
            anchorcolor = black]{hyperref}

\usepackage{pifont}
\usepackage{xspace}

\newcommand{\todoc}[2]{{\textcolor{#1}{\textbf{#2}}}}

\newcommand{\todored}[1]{{\todoc{red}{\textbf{[#1]}}}}

\newcommand{\todoblue}[1]{\todoc{blue}{\textbf{[#1]}}}

\newcommand{\minus}{\scalebox{0.75}[1.0]{$-$}}
\newcommand{\xz}[1]{\todored{XZ: #1}}

\newcommand{\ql}[1]{\todoblue{QL: #1}}

\newcommand{\ours}[1]{\textbf{Ours}}

\newcommand{\cmark}{\ding{51}}
\newcommand{\xmark}{\ding{55}}

\usepackage{bm}

\def\1{\bm{1}}

\def\rx{{\textnormal{x}}}
\def\ry{{\textnormal{y}}}

\DeclareMathAlphabet{\mathsfit}{\encodingdefault}{\sfdefault}{m}{sl}
\SetMathAlphabet{\mathsfit}{bold}{\encodingdefault}{\sfdefault}{bx}{n}

\def\gD{{\mathcal{D}}}

\def\gI{{\mathcal{I}}}

\def\gS{{\mathcal{S}}}

\DeclareMathOperator{\relu}{ReLU}

\newcommand{\system}{{\textsc{{D$^2$B}}}}
\newcommand{\systemspace}{{\textsc{{D$^2$B}}}\xspace}
\renewcommand{\citet}{\cite}

\begin{document}
\author{
 Qiuling Xu \\
  Department of Computer Science\\
  Purdue Univeristy\\
  \texttt{xu1230@purdue.edu} \\
   \And
 Guanhong Tao \\
  Department of Computer Science\\
  Purdue Univeristy\\
  \texttt{taog@purdue.edu} \\
  \And
 Xiangyu Zhang \\
  Department of Computer Science\\ Purdue Univeristy\\
  \texttt{xyzhang@cs.purdue.edu} \\
}

\date{}

\title{\Large \bf \systemspace: Deep Distribution Bound for Natural-looking Adversarial Attack}

\maketitle

%

\begin{abstract}
We propose a novel technique that can generate natural-looking adversarial examples by bounding the variations induced for internal activation values in deep layer(s), through a distribution quantile bound and a polynomial barrier loss function. 
By bounding model internals instead of individual input pixels, our attack admits perturbations closely coupled with the original input's existing features, allowing the generated examples to be natural-looking while having diverse and often substantial pixel distances from the original input. Enforcing per-neuron  distribution quantile bounds allows addressing the non-uniformity of internal activation values. 
Our evaluation on ImageNet and five different model architectures demonstrates that our attack is effective. Compared to five other state-of-the-art adversarial attacks in both the pixel space and the feature space, our attack can achieve the same success rate and confidence level while having much more natural-looking  adversarial perturbations. These perturbations piggy-back on existing local features and do not have any fixed pixel bounds. 


\end{abstract}
%

\section{Introduction}
\label{sec:introduction}

\begin{figure*}[ht]
    \centering
    \includegraphics[width=0.9\textwidth]{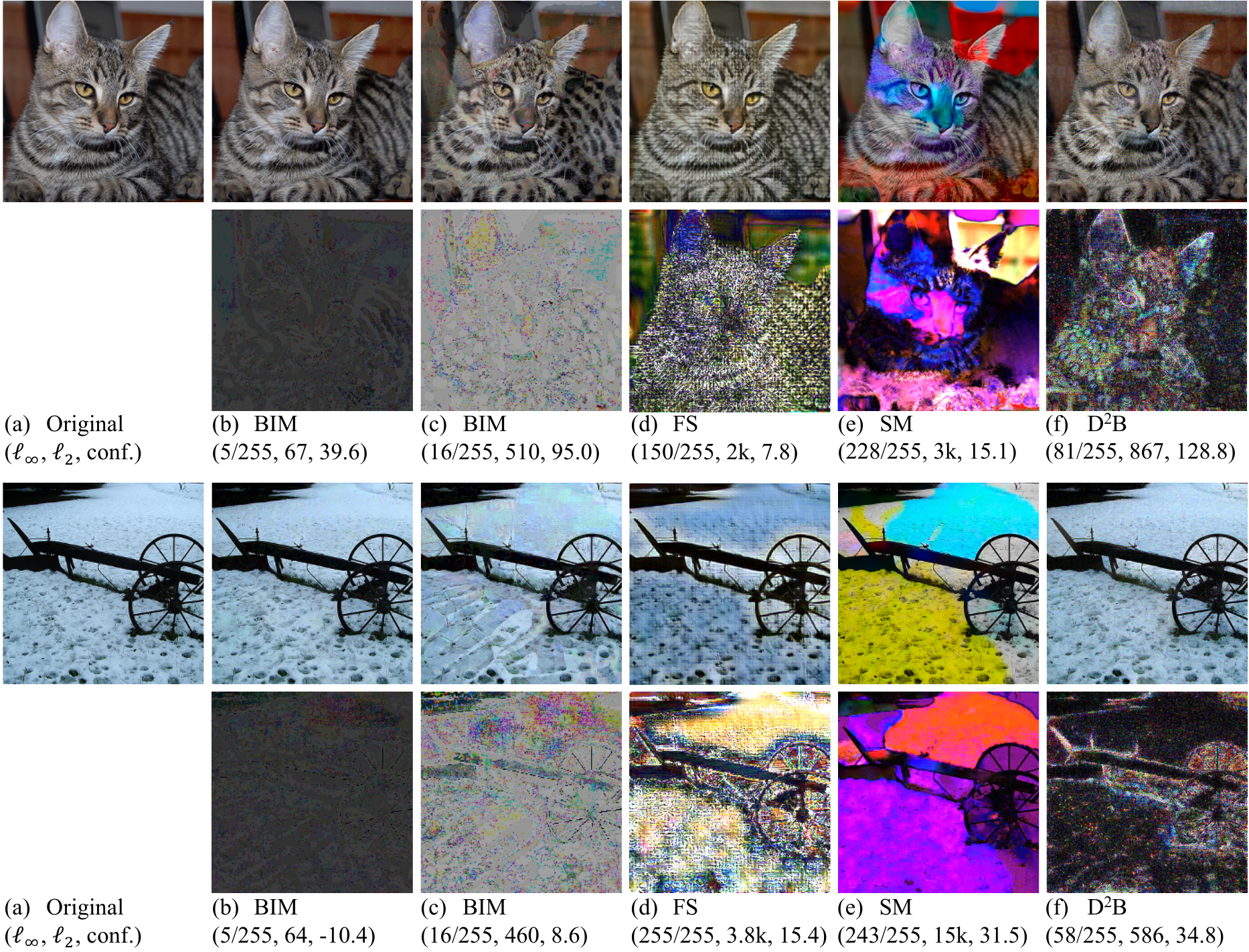}
    \caption{Adversarial examples of different attacks/attack-settings. The first column represents the original images. The second and third columns show examples from Basic Iterative Method (BIM) with a small and a large pixel distances, respectively. The following columns are samples from Feature Space Attack(FS), Semantic Attack (SM), and our \textbf{D}eep \textbf{D}istribution \textbf{B}ounded Attack (\system). For each set of images, the first row presents the adversarial examples. The second row shows the perturbations applied to the original image. We enlarge the perturbations of BIM by 10 times and the others' by 5 times for better illustration. On the bottom of each column, there is a triple representing the $\ell_\infty$, $\ell_2$ distances and the attack confidence of each example. A positive value indicates the attack yielded misclassification and a large value indicates the model is very confident about the (misclassification) result. 
    } 
    \label{fig:smallandlargepixelbound}
\end{figure*}

Adversarial attack is a prominent security threat for Deep Learning (DL) applications. With a benign input, perturbation is applied to the input to derive an adversarial example, which causes the DL model to misclassify. The perturbation is usually small, e.g.,  $[\minus 4 , 4]$ in the RGB range of $[0,255]$,
such that it is imperceptible by humans. Depending on the methods to generate such adversarial examples, there are {\em white-box} attacks, such as PGD~\cite{madry2017towards}, C\&W~\cite{carlini2017towards}, BIM~\cite{kurakin2016adversarial}, and FGSM~\cite{goodfellow2014explaining}, which assume access to model internals and leverage gradient information in sample generation; There are also {\em black-box} attacks, such as ZOO~\cite{chen2017zoo} and boundary attack~\cite{brendel2017decision}, that assume no access to model internals and  directly mutate inputs based on classification outputs. Our work falls into the {\em white-box} attack category in the image classification domain. The perturbation bounds are critical for adversarial attacks because a large bound usually implies high attack success rate but less natural-looking examples. The second and third columns of Figure~\ref{fig:smallandlargepixelbound} show some samples with a small bound (i.e., $\ell_\infty=5/255$, meaning the maximum pixel value change is 5 out of 255) and a larger bound (i.e., $\ell_\infty=16/255$) for the BIM attack\footnote{We use BIM instead of other pixel space attacks such as PGD because we found that (compared to BIM) the random initialization of PGD degrades imperceptibility at a non-trivial scale, in exchange for just a slightly higher success rate. Hence, we consider BIM a more compelling baseline as we stress imperceptibility.}
Observe that with the larger bound, the adversarial perturbation is detectable by human eyes. 
As such, it is often assumed that adversarial perturbation has a small bound. The success of a large number of existing defense, verification, analysis, and validation techniques~\cite{madry2017towards,kurakin2016adversarial,xu2018feature,xie2018mitigating,raghunathan2018certified,tramer2017ensemble,tao2018attacks,ma2019nic,katz2017reluplex,gehr2018ai2,wang2018formal,lecuyer2019certified,cohen2019certified,evasionatk, transferability}
are based on such assumptions. For example, given a particular input and a small bound, many techniques aim to verify/certify that a model does not misclassify for any perturbation within the bound~\cite{katz2017reluplex,gehr2018ai2,wang2018formal,lecuyer2018certified, ai2, rand_smooth, jia2019certified, wang2020certifying}. 

Researchers have recently shown that adversarial examples with large pixel distances (from the original inputs) can be generated. Such distances are usually way beyond the bounds that many existing defense and validation techniques aim to protect, providing a new attack vector. Specifically,
{\em semantic attack}~\cite{semantic_attack} manipulates a benign image's color and texture, through a specifically modified colorization model and a texture transfer method. 
{\em Feature space attack}~\cite{xu2020towards} leverages {\em style transfer}~\cite{huang2017arbitrary} to mutate the (implicit) styles of benign inputs to derive adversarial examples. In particular, it perturbs the distribution (e.g., mean and variation) of internal feature maps to inject malicious and largely human-imperceptible style differences that lead to misclassification.
The third and fourth columns in Figure~\ref{fig:smallandlargepixelbound} show some examples of feature attack and semantic attack, respectively. 
Observe that they have much larger $\ell_\infty$ and $\ell_2$ distances (from the original inputs in the first column) than the examples generated by BIM attacks. 
While they are more natural-looking than examples generated by pixel space attacks with a similar distance, the perturbations are quite noticeable in human eyes. This is confirmed by our human study (Section~\ref{sec:quality}), in which we show that with an 80\% attack success rate, humans can easily recognize the adversarial examples. The root cause is that these techniques focus on mutating {\em meta-features} of original inputs, such as colors and styles, due to the difficulty of harnessing perturbations on {\em content features}, such as shapes and local patterns. 
However, a successful semantic/feature attack may still entail substantial meta feature mutation, degrading the attack's  stealthiness. More discussion  on related work can be found in Section~\ref{sec:related_work}. 

In this paper, we propose a novel adversarial attack that can perform stealthy content feature mutation. The last column of Figure~\ref{fig:smallandlargepixelbound} shows the examples generated by our technique and its pixel-level contrast with the original image. Observe that the differences largely piggyback on the content features of the original example 
(by having similar shapes and local content patterns as the original inputs), making them human imperceptible. In comparison, the differences of the examples generated by semantic and feature space attacks are more pervasive and global. For example, the feature space attack on the cat image induces a global checkerboard style, while our attack induces perturbations that reside as part of the local features of the cat. The semantic attack on the barrow image below generates visible color blocks, while our attack induces more perturbations for places that have intensive local features and less perturbations for those with disperse local features such as the snowy background. 
According to our human study in Section~\ref{sec:quality}, our technique can achieve 95\% attack success rate, and yet humans cannot easily distinguish the adversarial examples from the benign ones. 

The essence of our approach is to bound internal activation changes instead of bounding the pixel space changes like many existing techniques.
Assuming internal neurons represent implicit features, particularly content features, small perturbations to the activations of these neurons denote small variations of the corresponding features. However, we find it  challenging to control internal perturbations properly. 
A naive method of limiting the variation of activation values after ReLU function to a range does not work and leads to abrupt pixel space perturbations that are human visible.
The reason is that a small activation change after ReLU may entail substantial value change before the ReLU and eventually human-perceivable pixel mutations in the input space. 
Moreover, while in the pixel space, a perturbation value range has consistent meaning/effect across multiple pixels, an internal value perturbation range does not possess such consistency across multiple neurons. 
For instance, small activation  changes may significantly affect classification results and entail substantial pixel space changes (in order to achieve such inner differences) for some neurons, while activation changes with an orders-of-magnitude larger scale may have very little effect for some other neurons. These pose new challenges to the underlying optimization methods.

To address these challenges,
we propose a novel solution as follows.
Given a model, its internal structure is inspected to select a so-called {\em throttle plane}, which  forms a complete cut for the data path from the input to the output and has values roughly following  normal distributions. 
It is a notion similar to, but with a finer granularity than, a layer in the deep learning model. Intuitively, consider that a layer consists of many operations, such as matrix multiplication, vector additions, and activation functions. A throttle plane 
consists of all the values produced by some such operation in a particular layer.
ReLU functions are not good candidates for the throttle plane as the activation value distributions for individual neurons (over all training samples) usually do not follow a normal distribution (see Section~\ref{sec:throttle_plane}).
After identifying the throttle plane, we collect its inner value distribution over the training set for each neuron on the plane. Such distributions allow us to provide {\em neuron-specific} bounds during perturbation. These bounds are defined based on the normal distributions (e.g., 10\% quantile change).
We then enforce the bound using  a {\em polynomial internal barrier loss} (Section~\ref{sec:poly_barrier_loss}).
The {\em barrier method} (BM) \cite{boyd2004convex} is a method developed for constrained optimization problems.
Intuitively, it adds a substantial penalty through a barrier loss when a value approaches its boundary.
By default, BM uses a log-based barrier loss, which is not ideal and has numerical instability problems in our context. We hence develop a new polynomial loss function. Our contributions are summarized as follows.

\begin{itemize}
    \item We develop a novel adversarial attack that perturbs content features (instead of meta features) so that the generated adversarial examples are imperceptible to humans.
    \item We identify that a direct extension of existing pixel space adversarial attack does not regulate internal perturbation well.
    \item We propose a novel distribution bound and a polynomial boundary loss function that enable effective internal regulation.
    \item We conduct experiments on ImageNet and five models, including both naturally and adversarially trained models. Comparative experiments with five state-of-the-art related attacks  
    show that our attack produces adversarial examples that are more natural-looking when achieving the same level of attack confidence/success-rate.
Further evaluation against three different detection techniques demonstrates that our attack has better/comparable persistence compared to other attacks while having  better imperceptibility, due to the new vulnerable aspects it attacks.

\end{itemize}

\begin{figure}
    \centering
    \includegraphics[width=0.49\textwidth]{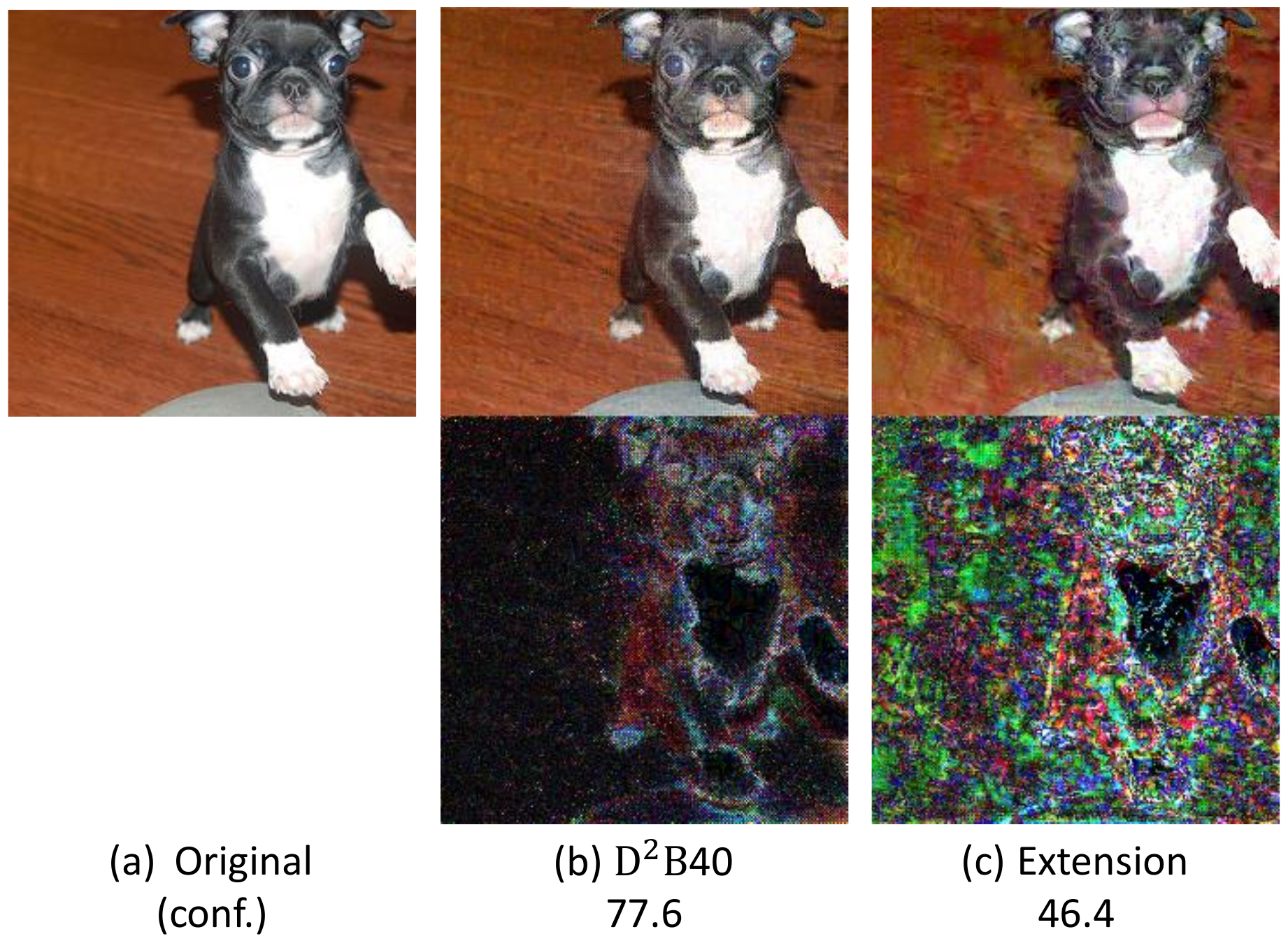}
    \caption{
    Untargeted Attack on ResNet152-Adv Using An Extension of Classic Pixel-space Attack and Our Attack. The second column is our method. For the extension, we use a 2\% min-max bound. See Section~\ref{sec:compare_diff_bound} and \ref{sec:opt_methods} for details. The first row presents the original image and the corresponding adversarial samples. The second row visualizes the perturbance applied on the natural image. Conf. means attack confidence. 
    }
    \label{fig:motivation_naive}
\end{figure}

\vspace{-5pt}
\section{Attack Design}
\label{sec:design}
\vspace{-5pt}


\begin{figure*}

    \centering
    \includegraphics[width=0.85\textwidth]{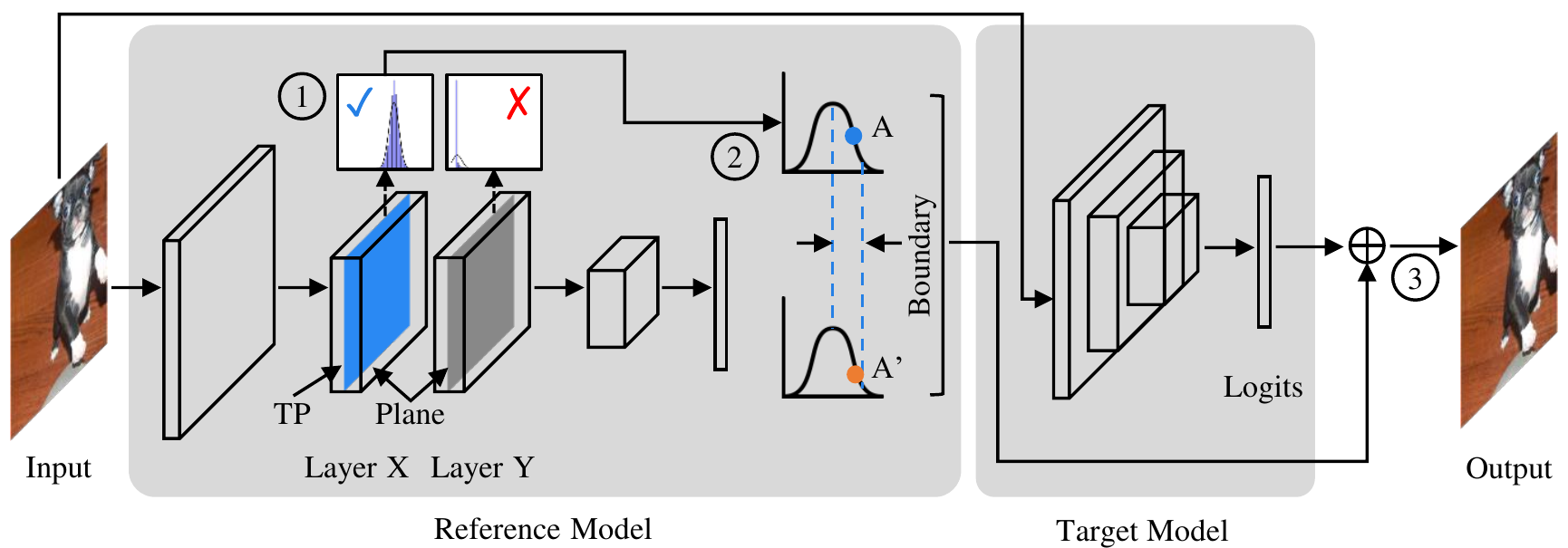}
    \vspace{-5pt}
    \caption{Workflow of our attack. It consists of three steps: \textcircled{1} throttle plane (TP) selection, \textcircled{2} internal distribution boundary constraint, and \textcircled{3} adversarial sample generation with combined losses.}
    \label{fig:workflow}
\end{figure*}

\noindent
{\bf Motivation - Direct Extension of Pixel Space Adversarial Attack Does Not Work Well.}
There is a large body of existing works that generate adversarial samples by constraining perturbation in the pixel space, such as PGD~\cite{madry2017towards}, C\&W~\cite{carlini2017towards}, BIM~\cite{kurakin2016adversarial}, and FGSM~\cite{goodfellow2014explaining}. They have a similar 
working mechanism. Take BIM as an example. Given a perturbation bound such as $8/255$, BIM iteratively perturbs individual input pixels following the gradient direction of a cross-entropy loss function (that tries to induce misclassification). When a perturbed pixel value exceeds the bound, it simply clips the value. The first design we explored was an extension of BIM in 
the feature space. Specifically, we first identify internal neuron activation value ranges by profiling on the training input. Like the value
bound in the pixel space attacks, we 
use a {\em min-max} bound that determines the lower and upper bounds 
of an internal activation value based on its range and a fixed percentage. For example, assume the profiled value range of a neuron is $[0,1000]$, and the current value of the neuron for a benign sample is 500, a 10\% bound allows the value to
vary in $[400,600]$. We clip the activation value if it exceeds the bound. Note that clipping suggests that the gradient of this neuron becomes 0, and hence it has no effect in updating input pixels. We update input 
pixels according to gradients just like in BIM. Instead of bounding the input pixel variations, we regulate the 
internal activation value variations.

However, we find that such a direct extension does not work well. Figure~\ref{fig:motivation_naive} shows a benign example, the adversarial version by our technique, and the adversarial version by the extension. Observe that the 
sample by the extension does not look natural and has a smaller confidence value compared to ours.
The reason is that internal space is 
different from the pixel space. A fixed (percentage) value range makes sense in the pixel space as it directly reflects a fixed level of human perception variation. In contrast, a fixed internal percentage value range may imply various levels of pixel changes and hence various human perception levels. A comparative experiment can be found in Section~\ref{sec:opt_methods}. This motivates our new attack design. $\Box$

\smallskip
\noindent
{\bf Our Design. }
\autoref{fig:workflow} describes the workflow of our attack. 
Given a reference model (from which a throttle plane is identified and used to regulate feature variations), a target model (for which adversarial samples are generated), and some training samples, we first perform throttle plane (TP) selection (step \textcircled{1}). 
Specifically, we run the reference model over the samples and collect the activation value 
distributions at the end of individual operations along the forward data path (e.g., the output of multiplication with a kernel matrix).   
{\em The values produced by 
 parallel operations in a particular layer across all neurons and channels form a plane} (e.g., the blue and gray planes in \autoref{fig:workflow}).
Intuitively, a plane is a ``vertical slice'' of a layer (and hence also a complete cut of the data path) and a layer can be considered as a stack of multiple planes. For instance, assume a layer consists of three operations: kernel multiplication, addition with bias, and ReLU activation function. The values collected at the end of each operation constitute a plane.

A plane whose value distribution has a normal distribution is a possible throttle plane (TP) to harness the adversarial perturbation (e.g., the blue plane in \autoref{fig:workflow}).
With a (or multiple) selected throttle plane(s), we further inspect the possible distribution boundary for each neuron at step \textcircled{2}. That is, the perturbed value $A'$ should be bounded within some distribution quantile range of the original value $A$. Finally, we model the constraint of distribution boundary by an internal barrier loss function (on the reference model), which is combined  with the cross-entropy prediction loss (on the target model). During attack (step \textcircled{3}), a normal input is fed to both models and updated with respect to the combined attack loss, which produces a successful natural-looking adversarial example.
While the reference model and the 
target model could be the same, 
empirically, we find that using a stand-alone reference model allows the best performance. The reason is that
depending on the model structure, a good  throttle plane may not exist in a target model. 


\vspace{-5pt}
\subsection{Distribution based Bounding and Throttle Plane Selection}
\label{sec:throttle_plane}


The overarching design of our attack is to harness perturbation at the selected {\em throttle plane(s)} 
such that only small variations of 
abstract features are allowed. Note that the corresponding pixel space perturbations could be substantial as long as the inner value changes are within bound.
Identifying appropriate throttle plane(s) is the first challenge we need to address. Intuitively, if we consider model execution as horizontal data flow from the input space to the output space, 
{\em a plane contains the values lie in a vertical cut of the data flow}. The cut could lie in the border between layers or even in between operations within a layer. 
Formally, {\em a plane consists of all the values right after parallel operations along the data flow from the input space to the output space across all the neurons/channels}. 
As such, the input 
values and the output values form planes; the activation values right after a layer's activation function form a plane too.

\begin{figure*}
    \centering
    \begin{subfigure}{.16\textwidth}
        \centering
        \includegraphics[width=\linewidth]{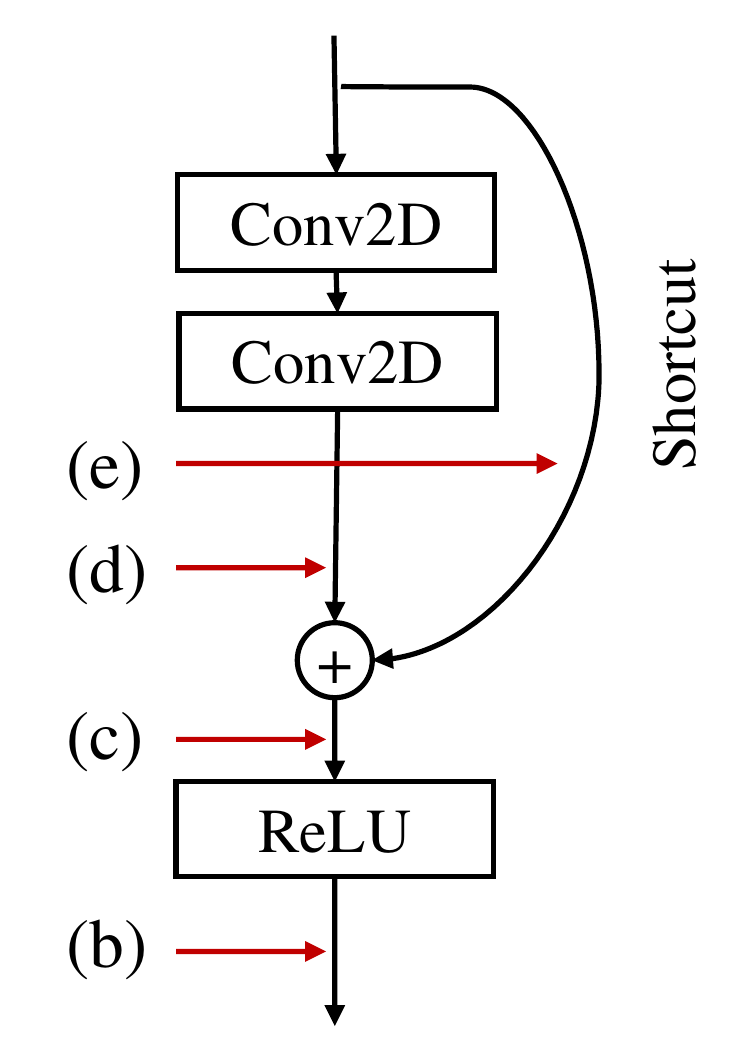}
        \caption{ResNet V1 Block}
        \label{fig:tpdist_position}
    \end{subfigure}
    \begin{subfigure}{.20\textwidth}
        \centering
        \includegraphics[width=\linewidth]{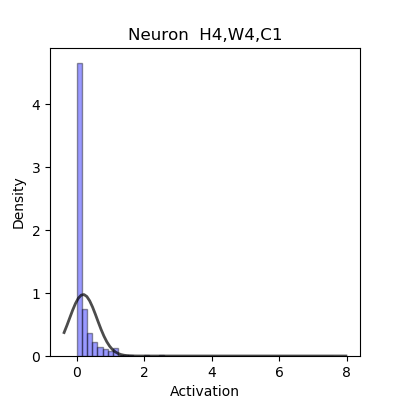}
        \caption{After ReLU}
        \label{fig:thdist_after_relu}
    \end{subfigure}
    \begin{subfigure}{.20\textwidth}
        \centering
        \includegraphics[width=\linewidth]{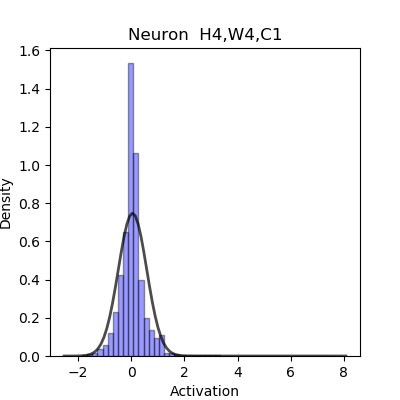}
        \caption{Before ReLU}
        \label{fig:thdist_before_relu}
    \end{subfigure}
    \begin{subfigure}{.20\textwidth}
        \centering
        \includegraphics[width=\linewidth]{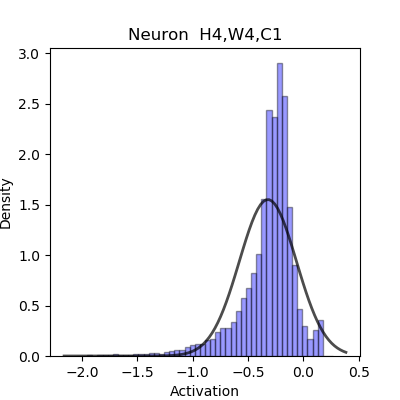}
        
        \caption{\small Main}
        \label{fig:thdist_nsc}
    \end{subfigure}
    \begin{subfigure}{.20\textwidth}
        \centering
        \includegraphics[width=\linewidth]{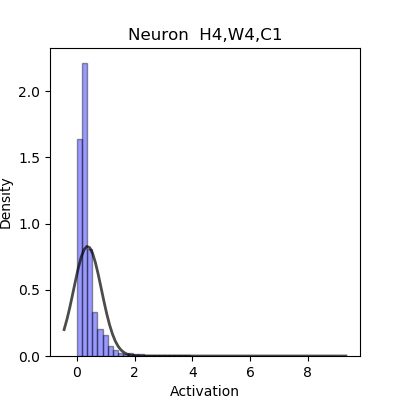}
        
        \caption{\small Shortcut}
        \label{fig:thdist_nsc}
    \end{subfigure}
    \vspace{-5pt}
    \caption{Operations in  the 
    last block of group 1 of an adversarially trained ResNet152 and the corresponding typical distributions of the values after these operations}
    \label{fig:throttle_distribution}
\end{figure*}

\begin{figure*}
    \centering
    \begin{subfigure}{.19\textwidth}
        \centering
        \includegraphics[width=\linewidth]{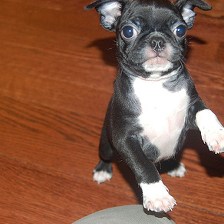}
        \caption{Original}
        \label{fig:th_img_real}
    \end{subfigure}
    \begin{subfigure}{.19\textwidth}
        \centering
        \includegraphics[width=\linewidth]{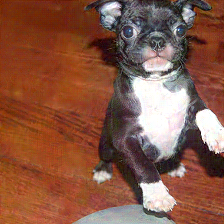}
        \caption{After ReLU}
        \label{fig:th_img_after_relu}
    \end{subfigure}
    \begin{subfigure}{.19\textwidth}
        \centering
        \includegraphics[width=\linewidth]{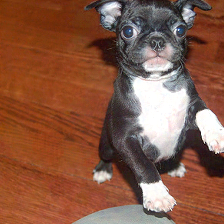}
        \caption{Before ReLU}
        \label{fig:th_img_before_relu}
    \end{subfigure}
    \begin{subfigure}{.19\textwidth}
        \centering
        \includegraphics[width=\linewidth]{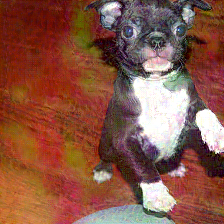}
        \caption{Main}
        \label{fig:th_img_nosc}
    \end{subfigure}
    \begin{subfigure}{.19\textwidth}
        \centering
        \includegraphics[width=\linewidth]{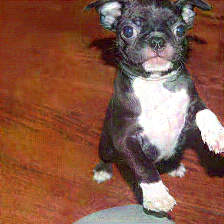}
        \caption{Main and Shortcut}
        \label{fig:th_img_nosc}
    \end{subfigure}
    \vspace{-5pt}
    \caption{Adversarial images when the throttle plane is at different positions in 
    Figure~\ref{fig:throttle_distribution},
    under the same perturbation bound. }
    \label{fig:throttle_image}
\end{figure*}

\noindent
{\bf Challenges of Having Internal Throttle Plane.}
 Traditional adversarial sample generation techniques simply place the perturbation throttle at the input plane. This makes the design simple as the perturbation happens strictly within the throttle plane.
 In contrast, placing the throttle in an inner plane poses new challenges.    

First of all, while in the input space
values have uniform semantics (e.g., denoting the RGB values of individual pixels), values in the inner space do not have such property. The different values
on the same inner plane often represent 
various abstract features whose value
ranges have diverse semantics. As such,
a uniform perturbation bound
across all these internal values is 
meaningless. Second, in our design, 
the perturbation occurs in the input
space while the throttle is placed somewhere inside the reference model. Hence, the input perturbation is not directly controlled and could be substantial. An important hypothesis
is that since the perturbations can only induce bounded inner value changes at the throttle plane, they denote small semantic mutations of the abstract features. However, given a particular inner value, the semantic
mutation entailed by its changes is non-uniform within its range. Consider Figure~\ref{fig:throttle_distribution}(c), which denotes the distribution of an inner value (across the training set). Observe that variation of 0.5 when the value is 1 implies much more substantial semantic changes (indicated by the entailed substantial quantile change) than when the value is 4, which is at the very tail of the distribution, as the model is likely insensitive to such a large value.

\noindent
{\bf Our Method -- Looking for Planes with Gaussian Distribution.} 
According to the above discussion, we cannot utilize a uniform bound across the different inner values (on the plane); we cannot utilize the same bound even when the value varies (from one input to another).  
Therefore, we propose a novel idea of 
using distribution based bounds. 
Particularly, we collect the distributions for the individual values
(on the plane). During perturbation,
the bound for each inner value is based
on its distribution. As such, not only
different values along the plane have
different bounds, but also, the value may have different bound when it varies from input to input.
In particular, we {\em select the plane(s)
whose values have Gaussian distributions} and {\em use a quantile bound based on the current value and its distribution}, instead of using a concrete value bound. These allow us to have precise and relatively easy control of the level of semantic mutation.
In the following, we use a few examples to illustrate the reasons for looking for an inner throttle plane with Gaussian distributions.


Take a block of an adversarially trained ResNet152 for example (Figure~\ref{fig:throttle_distribution} (a)).
If we set the throttle plane at the block boundary (i.e., right after the ReLU function), the distribution for some value on the plane (across the entire training set) is shown in Figure~\ref{fig:throttle_distribution} (b). Observe that while the distribution is dense on the positive side, the negative side is vacant (due to ReLU). It is hence not a good choice
for the throttle plane. The reason is that
we completely lose control on the negative side. Intuitively, substantial input perturbations would be admitted as long as they do not 
cause the inner value to flip from negative to positive after ReLU. This ignorance would degenerate the naturalness of the generated adversarial samples (see Figure~\ref{fig:throttle_image}(b) for an example).
If we set the plane right before ReLU, 
according to Figure~\ref{fig:throttle_distribution} (c), it is a Gaussian distribution. By inspecting the block structure, we find that the value is the sum of the shortcut and the main output (please refer to~\cite{he2016deep} for the explanation of these terms of ResNet structure). 
The Gaussian distribution makes enforcing a quantile bound easy and hence allows generating natural-looking adversarial examples if we place the throttle plane there (see Figure~\ref{fig:throttle_image}(c)).
Observe that the background has undergone much less perturbation (compared to others) as most perturbation is on the content features of the dog and hence not that visible.
We also study the distributions of 
the values after the main output and along the shortcut 
(e.g., Figure~\ref{fig:throttle_distribution}(d) and (e), respectively). 
Observe that although (d) denotes a rough Gaussian distribution, placing the throttle there produces unnatural samples (see Figure~\ref{fig:throttle_image}(d)). 
The main reason is that the operation alone does not constitute a plane as it is not a complete cut of the data path.
Intuitively, it leaves a large part (i.e., the values along the shortcut) unregulated. 
Since the operations (d) and (e) together form a cut, we also study placing the throttle at these two operations simultaneously. However, the results (see Figure~\ref{fig:throttle_image}(e)) are also inferior to placing the throttle on a single operation (c). 
The reason is that while the changes 
on the two respective operations respect their bounds, their aggregation may not respect the bound. In contrast, placing 
the throttle right after they are aggregated has the best regulation. 

Figure \ref{fig:appendix_renset152_dist}(e) in Appendix~\ref{appendix: throttledist} shows the distributions for a set of randomly selected values on the same plane. Observe they approximately follow normal distributions. Also, observe that their distribution parameters are quite different, supporting our design of using different bounds for various values on a plane. 



During sample generation, given a benign input, the selected throttle plane's inner values are collected. The bound of the value is then determined by its quantile of the value (on its density function). How to enforce such quantile bounds is discussed in the following section.

\subsection{Enforcing Quantile Bound with Polynomial Barrier Loss}
\label{sec:poly_barrier_loss}

Let $\gD$ be a distribution on support $\gS$. The activation $\ry_i$ of neuron $i$ on a selected throttle plane $\gI$ is a random variable through mapping $f_i: \gS \rightarrow \mathbb{R}$, $\ry_i \sim f_i(\gD)$. We denote the cumulative distribution function of $\ry_i$ as $C_i(x)$, and the corresponding quantile function as $C_i^{-1}(x)$. Let the original image be $\rx^{\mathrm{nat}}$ and the adversarial sample be $\rx^{\mathrm{adv}}$. Correspondingly, let ${\ry_i}^{\mathrm{adv}}$ and ${\ry_i}^{\mathrm{nat}}$ be the respective activations for $\rx^{\mathrm{nat}}$ and $\rx^{\mathrm{adv}}$. Assume the allowed quantile change is less than a threshold $\epsilon$. The corresponding
value bound for $\ry_i^{\mathrm{adv}} \in [\mathrm{low}_i,\mathrm{high}_i]$, is hence the following.
\begin{align}
\begin{split}
     \ry_i^{\mathrm{adv}} \in & \left[C_i^{-1}\left(\max(C_i(\rx^{\mathrm{nat}})-\epsilon,0)\right), \right.  \left. C_i^{-1}\left(\min(C_i(\rx^{\mathrm{nat}})+\epsilon,1)\right)\right] \\
\end{split}
\label{eq:quantilebound}
\end{align}
 Note that we translate the quantile bound to a value bound, over which we can define a loss function.

\noindent
{\bf Polynomial Barrier Loss.}
{\em Interior point method} or {\em barrier method}~\cite{boyd2004convex}
is a standard technique for
constrained optimization. It is widely used in linear programming applications \cite{vanderbei2015linear}.
It utilizes a negative
log function in the loss function by default.
However, it was intended to be
used in problems where the bound is hard, meaning the values must not exceed the bound as the loss becomes infinitely large when the value infinitely approaches the bounds. In our context, a hard bound does not work well 
with ReLU functions. 
Specifically, input changes guided by gradients may activate some previously inactive neurons, leading to the inner values to exceed their bounds, causing numerical exceptions (on the log function).
Another naive design is to introduce a ReLU kind of bound, that is, the loss is 0 while the value is in bound and some large value otherwise. 
However, such a design does not apply penalty when the value is approaching the bound.

Therefore, we devise a polynomial barrier loss function as follows. 
\begin{align}
\mathcal{L}_i({\ry_i}^{\mathrm{adv}}) = k\left[\frac{\relu({\ry_i}^{\mathrm{adv}}-{\ry_i}^{\mathrm{nat}})}{\mathrm{high}_i - {\ry_i}^{\mathrm{nat}}}+ \frac{\relu({\ry_i}^{\mathrm{nat}}-{\ry_i}^{\mathrm{adv}})}{{\ry_i}^{\mathrm{nat}}- \mathrm{low}_i }\right]^{b}
\end{align}
We empirically set $k=1e5$ and $b=200$.
Intuitively, the loss function applies an extremely large penalty (by the power $b=200$) when the inner value induced by adversarial perturbation is beyond the bound. When the value is within the bound and close to the boundary values, 
a large penalty is applied. These penalties discourage the value from going beyond. For example, when
${\ry_i}^{\mathrm{adv}}$ is larger than ${\ry_i}^{\mathrm{nat}}$ and close to the upper bound $\mathrm{high}$, say 
$\frac{\relu({\ry_i}^{\mathrm{adv}}-{\ry_i}^{\mathrm{nat}})}{\mathrm{high} - {\ry_i}^{\mathrm{nat}}}=0.99$, the loss is $1e5 \times 0.99^{200}\approx 1e4$. We have also tried a linear barrier loss, which cannot effectively enforce the bound. Please refer to Section~\ref{sec:two_barrier_loss} 

\begin{figure*}[ht]

        \centering
        \includegraphics[width=.99\linewidth]{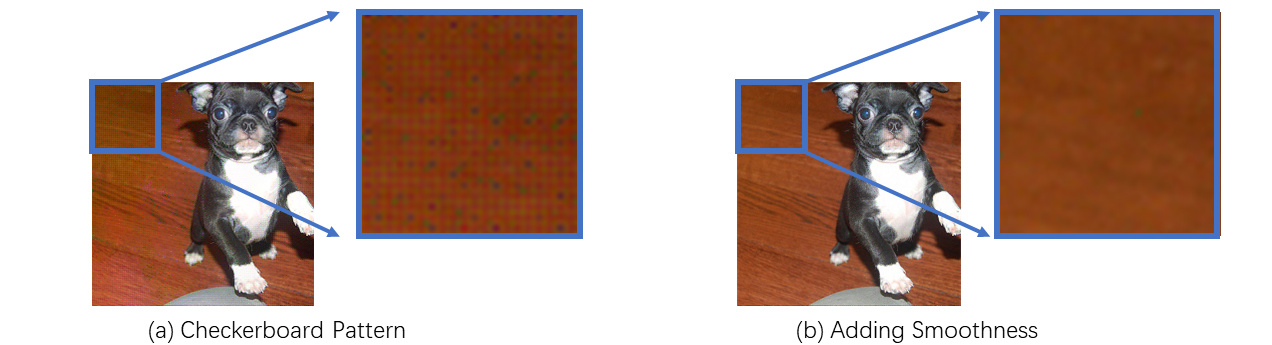}
        \caption{Results before and after smoothing for a VGG16 model}
        \label{fig:checkerboardpattern}

\end{figure*}

\smallskip
\noindent
{\bf Optimization Method.}
With the polynomial barrier loss, we use a standard gradient sign method \cite{carlini2017towards} for optimization. There are other design choices. For example, in ~\cite{DBLP:conf/ijcai/Kumari0SMKB19}, a two-step optimization was proposed to facilitate adversarial example generation by leveraging internal values. Specifically, it first perturbs the internal values at some inner layer to induce misclassification. It then uses a mean squared error loss to optimize the input to achieve the optimized inner values. However, we found that the method is not effective 
when a strict internal boundary is enforced. The reason is that the first step of inner layer optimization tends to find local minimus that is infeasible for input optimization due to the strong correlations across inner values. In contrast, our method directly optimizes the input space. 
Another simple method is clipping, which clips the inner values (on a throttle plane) and prevents gradient propagation if they are beyond bounds as we mentioned in Section~\ref{sec:design}.
Our experience shows that such  a simple method can hardly work either. 
We conduct experiments to compare the three methods. Our method can better enforce the internal bound and generate adversarial examples with one order of magnitude smaller average boundary size. Details can be found in Appendix~\ref{sec:opt_methods}. 
Identifying an appropriate quantile bound value is important. 
We address the problem by profiling the  quantile changes at the throttle plane under other attacks.  Specifically, we use the average observed internal value $\ell_\infty$ quantile change (at a throttle plane) for the adversarial examples by BIM4.
How to identify the appropriate learning rate is discussed in Appendix~\ref{sec:binarysearch}. 

\subsection{Feature Smoothing}
\label{sec:feature_smooth}
Occasionally, we observe the generated adversarial examples exhibit the checkerboard patterns. \autoref{fig:checkerboardpattern}(a) shows a typical adversarial example with checkerboard pattern (zoomed in on the right). We observe these cases often occur when we use
VGG16 as the reference model (the other reference model we use is ResNet152-Adv).
We speculate that this is because we enforce bounds for individual values (on a throttle plane) independently and do not consider their joint distribution of nearby neurons.
To mitigate the problem, we add a feature smoothing loss to the optimization goal. 
The intuition is that individual values (on a plane) have a similar trend of change with their neighboring values. Thus we calculate the average of surrounding changes and use a mean squared error loss to prevent the change from being too far away from the average.
Suppose $\ry \in \mathbb{R}^{D \times H\times W}$ denotes a  throttle  plane with channel $D$, height $H$ and width $W$. The quantile changes are written as $\Delta Q_{\mathrm{d,h,w}}=|C_{\mathrm{d,h,w}}(\ry_{\mathrm{d,h,w}}^{\mathrm{adv}})-C_{\mathrm{d,h,w}}(\ry_{\mathrm{d,h,w}}^{\mathrm{nat}})| / \epsilon$. 
The average of changes made to nearby values can be formulated with an average pooling operation  $\mathrm{AvgPool}_{3\times3}(\Delta Q)$. Thus we expect the smoothness loss, written as $\frac{\alpha}{D \cdot H \cdot W} \sum_{\mathrm{d,h,w}}[\mathrm{AvgPool}_{3\times3}(\Delta Q)_{\mathrm{d,h,w}} - \Delta Q_{\mathrm{d,h,w}}]^2$ to be small. We empirically set the weight of smooth loss to $\alpha=10$. This can lead to 6\% improvement in the human preference rate when 
VGG16 is used as the reference model. The smoothness loss largely eliminates the checkerboard pattern as shown in \autoref{fig:checkerboardpattern}(b).

\section{Experiments}
\label{sec:experiments}


\begin{figure*}[!h]
    \centering
    \begin{subfigure}{.49\textwidth}
        \centering
        \includegraphics[width=.99\linewidth]{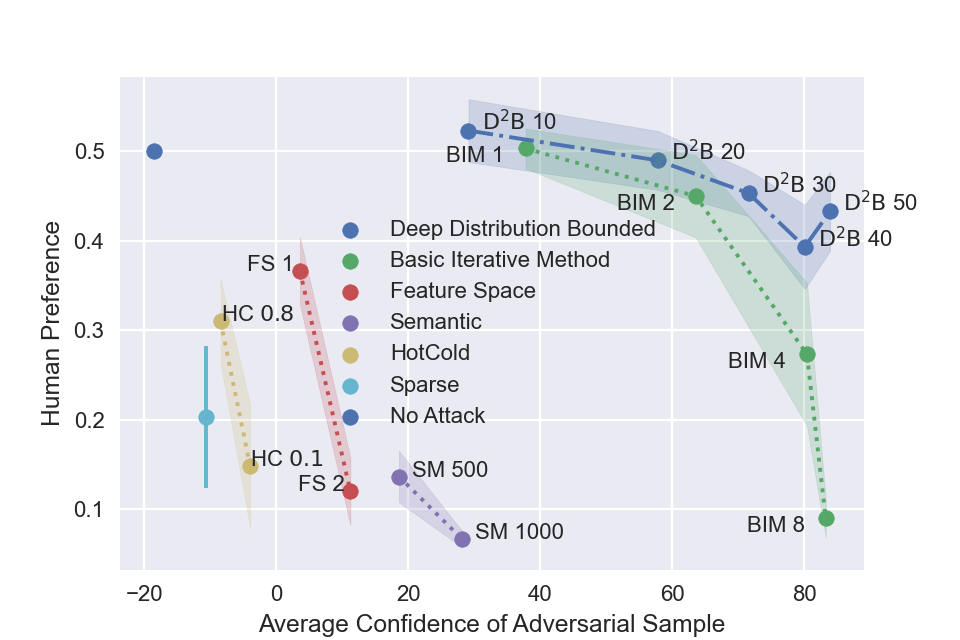}
        \caption{Targeted attacks on Resnet50}
        \label{fig:pref_resnet50}
    \end{subfigure}
    \begin{subfigure}{.49\textwidth}
        \centering
        \includegraphics[width=.99\linewidth]{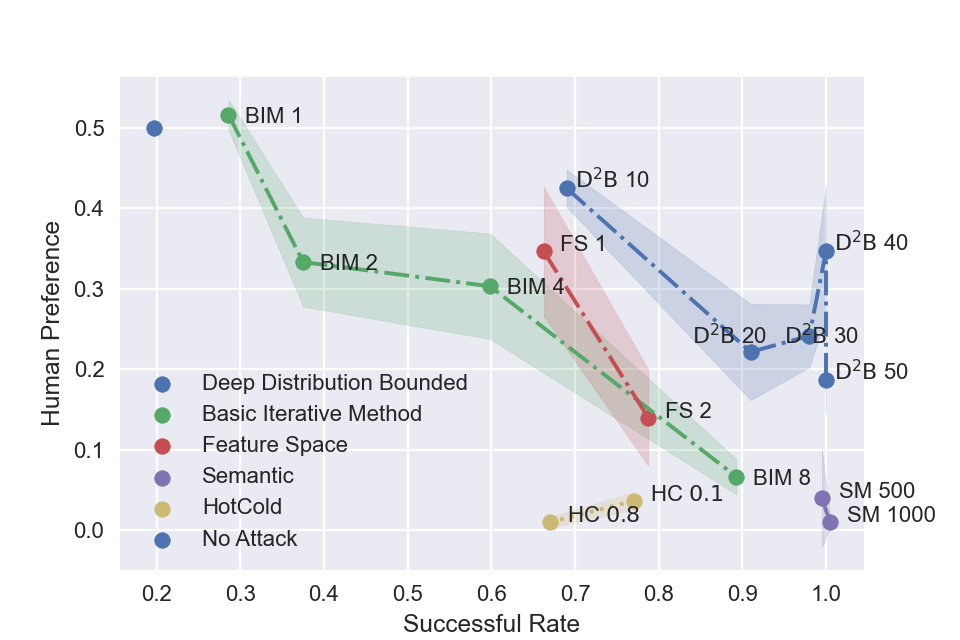}
        \caption{Untargeted attacks on Resnet152-Adv}
        \label{fig:pref_resnet152}
    \end{subfigure}

    \begin{subfigure}{.49\textwidth}
        \centering
        \includegraphics[width=.99\linewidth]{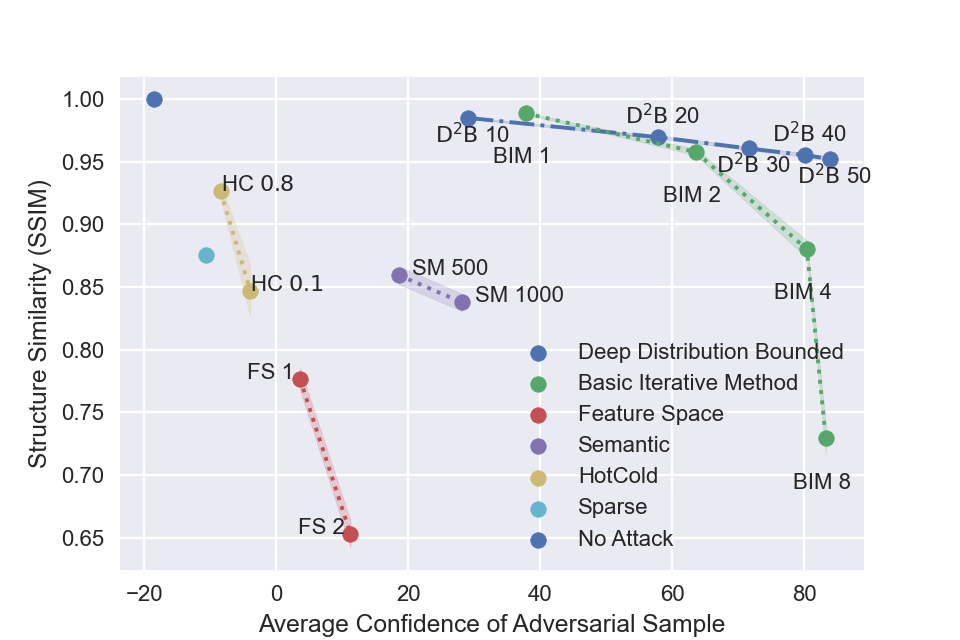}
        \caption{Targeted attacks on Resnet50}
        \label{fig:ssim_resnet50}
    \end{subfigure}
    \begin{subfigure}{.49\textwidth}
        \centering
        \includegraphics[width=.99\linewidth]{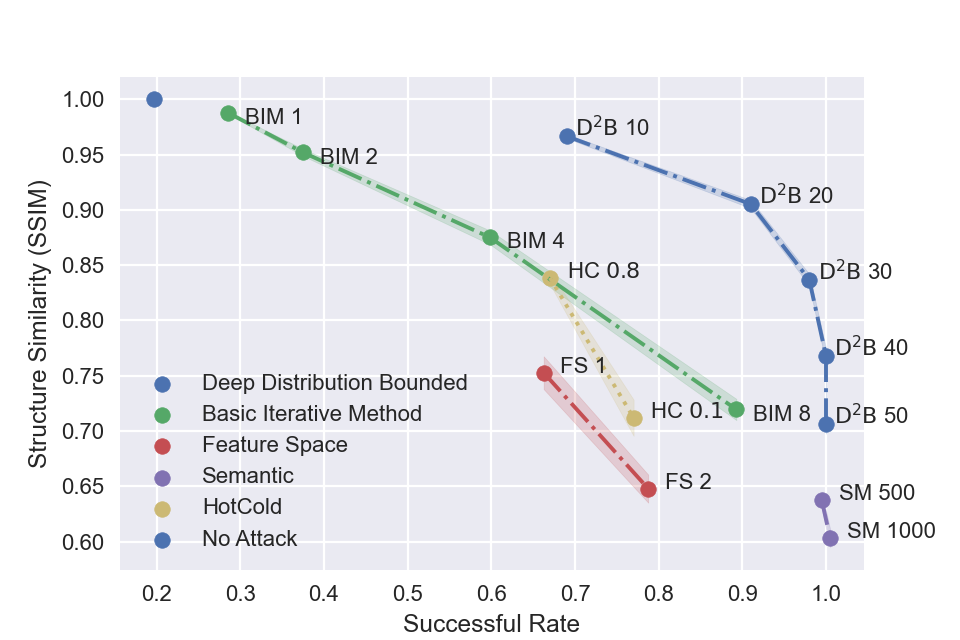}
        \caption{Untargeted attacks on Resnet152-Adv}
        \label{fig:ssim_resnet152}
    \end{subfigure}

    \caption{Quality of the generated adversarial examples. In (a) and (c) the targeted attacks,  the $x$-axis represents the average confidence of adversarial examples, where a confidence value indicates the level of success of the attack with a greater than 0 value meaning the model misclassifies to the target label. 
    In (a) and (b) the $y$-axis denotes the rate that humans consider the adversarial examples real (compared to the benign images), with 50\% meaning humans cannot distinguish an adversarial example from its benign version. In (c) and (d), the $y$-axis denotes the SSIM score indicating the similarity between images. A value 1 indicates they are identical. In (b) and (d) the untargeted attacks against the adversarially trained ResNet152, we use the attack success rate as the $x$-axis. Note that we cannot use attack success rate in the targeted attacks as all those attacks have close to 100\% success because the model was normally trained (and hence not robust). 
    We regard an untargeted attack successful if the true label does not appear in the top-5 predicted labels, which is consistent with the literature~\cite{Xie_2019_CVPR}. The shaded area on a data point (i.e., an attack setting) represents the standard error of the human preference rate or the SSIM score. In all these figures, a good method should be at the top-right corner.} 
    \label{fig:humanprefvssucc}
\end{figure*}

To evaluate our attack and the quality of generated adversarial examples, we conduct experiments on one of the largest image datasets, ImageNet~\cite{russakovsky2015imagenet}. We use five types of DNN models in the evaluation and compare the quality of our generated adversarial examples with five existing attack methods.  Finally, we evaluate our attack on three popular adversarial detection approaches.

\subsection{Visual Quality of Generated Examples}
\label{sec:quality}

In this section, we evaluate the visual quality of generated adversarial examples by our attack. We also compare our technique with five existing attack methods: BIM~\cite{kurakin2016adversarial}, hot cold attack~\cite{hotcold}, sparse attack~\cite{sparse}, feature space attack~\cite{xu2020towards} and semantic attack~\cite{semantic_attack}.
We use BIM as the representative of classic pixel space adversarial attacks such as PDG and C\&W (due to the reason explained in Section~\ref{sec:introduction}).
Feature space attack uses auto-encoder based on VGG16 and performs bounded perturbation of the mean and variance of internal embeddings~\cite{xu2020towards}. 
Semantic attack manipulates input color and texture.
Sparse attack applies small perturbation to salient pixels and large perturbation to less salient pixels. Hot-cold attack uses the SSIM score~\cite{SSIM} 
to constrain perturbation.
We use these six attack methods to generate adversarial examples for a naturally trained ResNet50 model~\cite{DBLP:journals/corr/HeZRS15} and an adversarially trained ResNet152 model~\cite{xie2019feature} (ResNet152-adv). For BIM, sparse attack, hot cold attack, feature space attack and our attack, we stop the attack optimization when convergence is reached (no confidence increase). For semantic attack, we use a preset number of optimization steps. 
Note that it is unbounded and the optimization  step controls the perturbation and the attack success rate. Note that for the sparse attack, we are unable to scale it up to an untargeted attack on ResNet152-Adv. 
Details about the throttle planes used are discussed in 
Appendix~\ref{sec:choose_throttle_planes}. 




\begin{table*}[ht]
\begin{minipage}[b]{0.48\textwidth}
\scriptsize
\centering
\tabcolsep=4.5pt
\caption{Pixel and quantile distances for ResNet50, with Conf. meaning attack confidence
}
\begin{tabular}{ccccccc}
\toprule
\multirow{2}{*}{Attack} & \multirow{2}{*}{Conf.} & \multicolumn{2}{c}{Pixel Dist.} & \multicolumn{3}{c}{$\ell_\infty$ Quantile Dist.} \\ \cmidrule(lr){3-4} \cmidrule(lr){5-7}
                        &                             & $\ell_2$               & $\ell_\infty$            & Plane 1      & Plane 2      & Plane 3      \\
\midrule                    
BIM4                    & 81.91                       & 10.81            & 0.04            & 0.62         & 0.83         & 0.90         \\
\system10                   & 30.29                       & 4.26             & 0.07            & 0.06         & 0.08         & 0.09         \\
\system20                   & 58.82                       & 6.27             & 0.09            & 0.12         & 0.16         & 0.17         \\
\system30                   & 72.43                       & 7.28             & 0.10            & 0.18         & 0.24         & 0.26         \\
\system40                   & 80.06                       & 7.88             & 0.11            & 0.24         & 0.32         & 0.35         \\
\system50                   & 84.52                       & 8.25             & 0.11            & 0.30         & 0.41         & 0.44 \\
\bottomrule
\end{tabular}
\label{tab:distance_resnet50}
\end{minipage}
\hfill
\begin{minipage}[b]{0.48\textwidth}
\scriptsize
\centering
\tabcolsep=4.5pt
\caption{Pixel and quantile distances for ResNet152-Adv, with Succ. meaning attack success rate}
\begin{tabular}{ccccccc}
\toprule
\multirow{2}{*}{Attack} & \multirow{2}{*}{Succ.} & \multicolumn{2}{c}{Pixel Dist.} & \multicolumn{3}{c}{$\ell_\infty$ Quantile Dist.} \\ \cmidrule(lr){3-4} \cmidrule(lr){5-7}
                        &                               & $\ell_2$               & $\ell_\infty$            & Plane 1      & Plane 2      & Plane 3      \\
\midrule                        
BIM4                    & 0.58                          & 14.49            & 0.04            & 0.65         & 0.82         & 0.89         \\
\system10                   & 0.61                          & 6.65             & 0.11            & 0.06         & 0.08         & 0.09         \\
\system20                   & 0.86                          & 15.43            & 0.24            & 0.13         & 0.16         & 0.17         \\
\system30                   & 0.86                          & 16.68            & 0.19            & 0.19         & 0.24         & 0.26         \\
\system40                   & 0.97                          & 21.84            & 0.22            & 0.26         & 0.33         & 0.35         \\
\system50                   & 0.98                          & 27.08            & 0.26            & 0.33         & 0.41         & 0.44         \\
\bottomrule
\end{tabular}
\label{tab:distance_resnet152adv}
\end{minipage}
\end{table*}

In order to measure the naturalness of generated examples, we perform a human study using Amazon Turk. We employ a similar setting as that in~\cite{zhang2016colorful}. Specifically, for each attack setting, users are given 100 pairs of images, each consisting of a real image and its adversarial counterpart. They are asked to choose the one that looks real. Each user is given 5 test-drives before the study starts. Each pair of image appears on screen for 5 seconds and is evaluated by three different users. 
The experiment is performed for each attack setting. 
There are 41 
settings (for the six
attacks, the comparison in Section~\ref{sec:compar_ref_models} and Section~\ref{sec:compare_diff_bound}), and 246 users participated in our study. 243 out of the 246 responses are considered valid, with those deviating far from the majority (exceeding two times of the standard deviation) removed. 
We post all the examples used in the human study on an anonymous website \cite{d2b2020online}. In addition to the human study, we also use Structure Similarity Index (SSIM)~\cite{SSIM} to quantify the perceptual distance of the adversarial samples. SSIM ranges from $-1$ to $1$, with a larger value indicating more similarity, while a 0 score indicating no similarity.



\autoref{fig:humanprefvssucc} shows the results. BIM$x$ denotes BIM with an $\ell_\infty$ bound  $x\%$ of 255. For example, BIM4 means the $\ell_\infty$ bound is  $255\times4\%\approx 10$. 
FS1 and FS2 are feature space attacks using the relu2\_1 and relu3\_1 layers of VGG16, respectively, as the embedding layer. SM$x$ is semantic attack with an optimization step of $x$ and the number of clusters set to 8~\cite{semantic_attack}. 
HC$x$ is the hot-cold attack using a SSIM score $x$ as the constraint. For example, HC $0.8$ admits adversarial samples with a SSIM score greater than $0.8$.  \system$x$
denotes that we allow $x\%$ of the average quantile change observed in BIM4 at the throttle plane. The reason we use percentage relative to BIM4 is for simplicity. Otherwise, one needs to fine tune the magnitude among different throttle planes. The confidence score of a sample $x$ is defined on the logits (the pre-softmax) value $L(x)$
\cite{carlini2017towards}. Specifically, for a targeted attack, supposing the target label is $t$, the confidence score of a sample $x$ is defined as follows.
\begin{align}
    L_t(x) - \max_{i\neq t} L_i(x)
    \label{eq:conf_target}
\end{align} Intuitively, it is the logits gap between the target label and another label with the maximum logits. 
For the untargeted attack, we evaluate the confidence over the top-5 classified labels. This is a common practice as in \cite{xie2019feature}. Suppose $l$ is the correct label and the $\mathrm{top}^i_{j\neq l} L_j(x)$ represents the $i$-th largest logit value other than the correct label $l$. The confidence is thus defined as follows.
\begin{align}
      \sum_{i=1}^{5} [ L_l -\mathrm{top}^i_{j\neq l} L_j(x)]
    \label{eq:conf_untarget}
\end{align}

\begin{figure*}[h]
    \centering
    \begin{subfigure}{.49\textwidth}
        \centering
        \includegraphics[width=.99\linewidth]{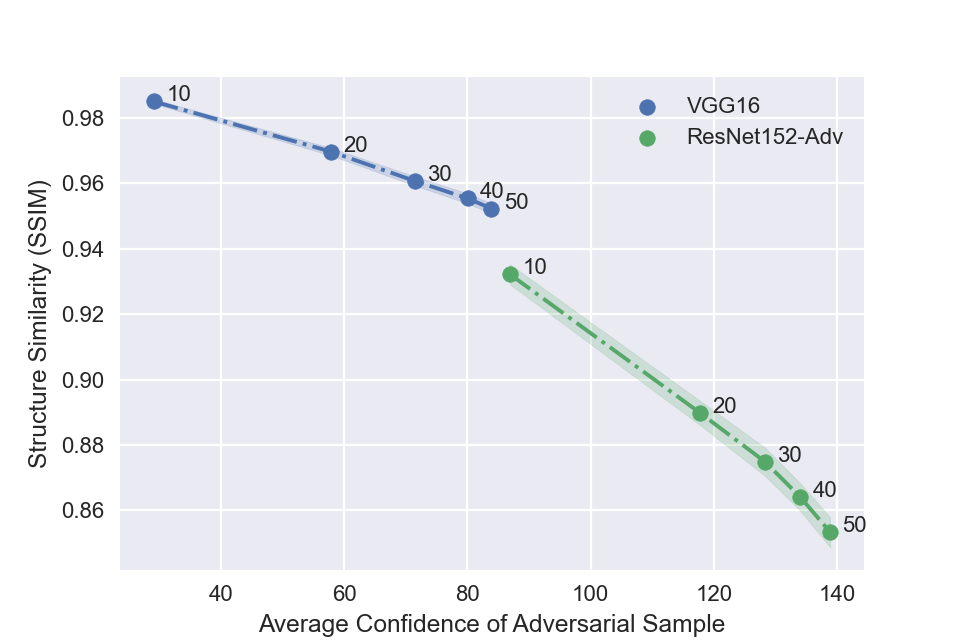}
    \end{subfigure}
    \begin{subfigure}{.49\textwidth}
        \centering
        \includegraphics[width=.99\linewidth]{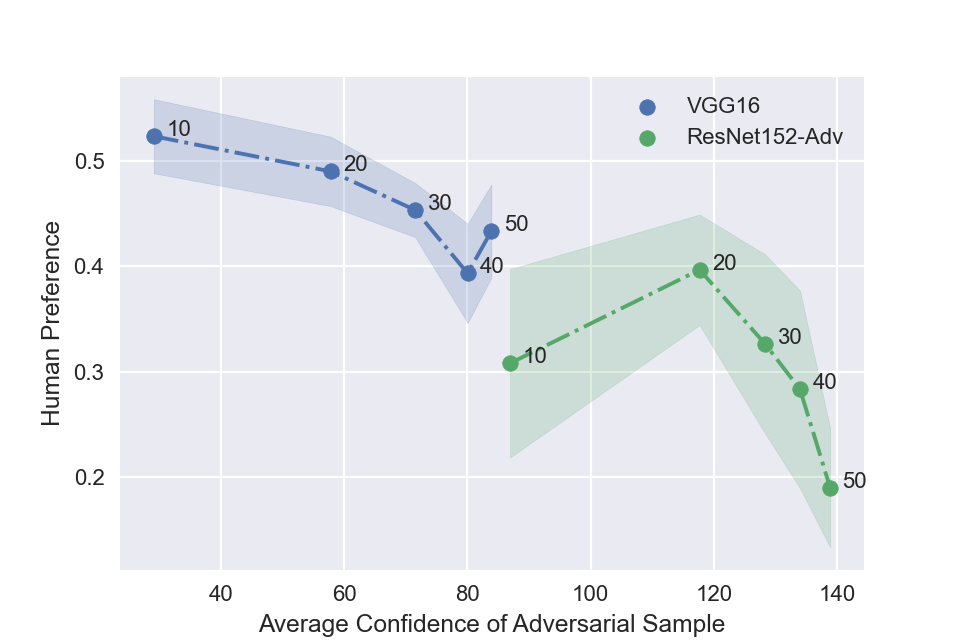}
    \end{subfigure}
    \caption{Attack Resnet-50 with different reference models. All the other settings are the same. 
    The exact locations of throttle planes can be found at Appendix \ref{sec:choose_throttle_planes}. }
    \label{fig:compare_ref_model}
\end{figure*}

\begin{figure*}[ht]
    \centering
    \begin{subfigure}{.49\textwidth}
        \centering
        \includegraphics[width=.99\linewidth]{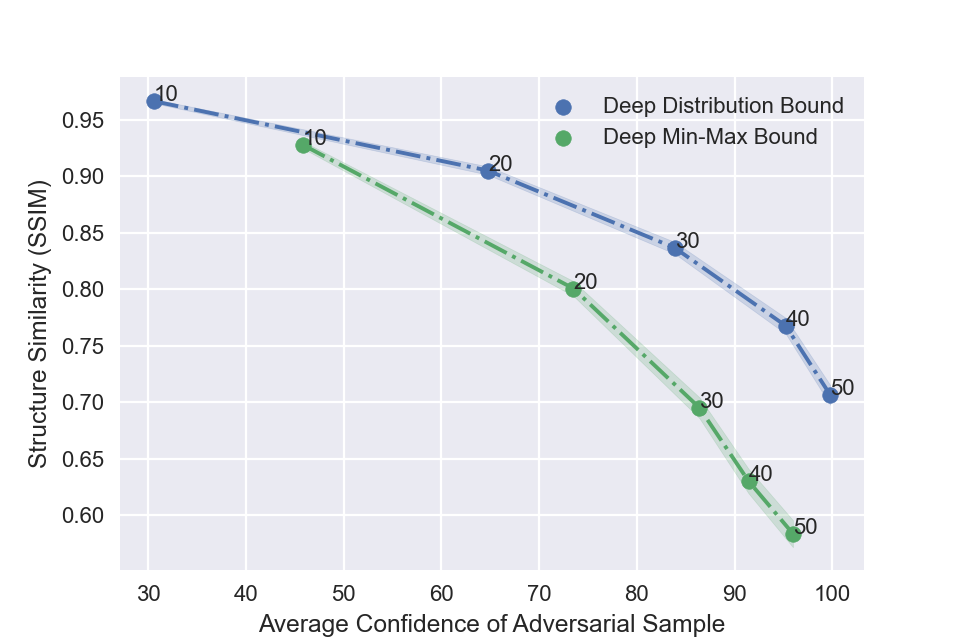}
    \end{subfigure}
    \begin{subfigure}{.49\textwidth}
        \centering
        \includegraphics[width=.99\linewidth]{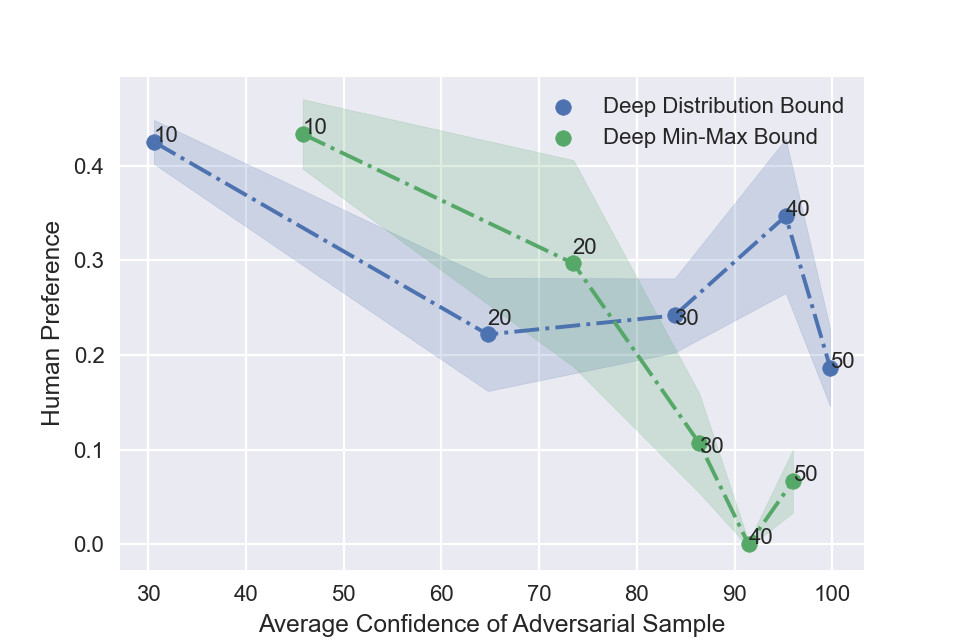}
    \end{subfigure}
    \caption{Comparison of Deep Distribution Bound and Deep Min-max Bound. We use the two bounds to attack Resnet152-Adv and report their human preference rate and SSIM Score. The annotated numbers represent the percentage of the bound relative to BIM4.
    }
    \label{fig:SSIM_Human_Bound_Type}
\end{figure*}

Intuitively, the score measures the success level of an attack.
Figures~\ref{fig:pref_resnet50} and \ref{fig:ssim_resnet50} show the results for targeted attacks on the naturally trained ResNet50. Figures~\ref{fig:pref_resnet152} and \ref{fig:ssim_resnet152} are for untargeted attacks on the adversarially trained ResNet152. We can see that our attack has the highest attack confidence/success rate at the same level of human preference/SSIM score. And with the same confidence/success rate, our adversarial examples are consistently more favored by the testers/SSIM score (for being more natural-looking) than those by other attacks.
Our adversarial examples with the most aggressive settings (e.g., \system40 and \system50) have a similar human preference to pixel space attack with a very small bound (BIM2 and BIM4), indicating our attack is indeed imperceptible. With the increase of quantile change, perturbation bound, or optimization step, all the attacks achieve a higher success rate, and our attack is increasingly more imperceptible than others. We further study the pixel distance and quantile distance of the generated adversarial examples by different attacks. 
The $\ell_\infty$ quantile distance is defined as $\max_{i \in \gI}|C_i(\rx^{\mathrm{adv}})- C_i(\rx^{\mathrm{nat}})|$.  \autoref{tab:distance_resnet50} and \autoref{tab:distance_resnet152adv} show that with a similar level of attack confidence or attack success rate, our attack has a smaller $\ell_2$ pixel distance and $\ell_\infty$ quantile distance. This indicates that our generated examples can achieve a similar level of attack effectiveness with less perturbation, they are more natural-looking. In other words, it can tolerate more aggressive perturbation without degrading naturalness as much, demonstrating the benefits of bounding deep layers.
The larger $\ell_\infty$ pixel distance and the smaller $\ell_2$ pixel distance (compared to BIM4) indicate our perturbations are more diverse, heavily piggy-backing on original features. The attack effectiveness and pixel/internal distances for other models are similar. Details can be found in Appendix~\ref{sec:more_models_dist}. More adversarial examples generated by the different settings of our attack and other attacks can be found in \autoref{fig:advscales_resnet50} and \autoref{fig:advscales_resnet152} in Appendix~\ref{appendix: adv_samples_scales}.
We also conduct a study about the essence of \system{} by studying the places that it aims to attack. Details can be found in 
 Section~\ref{sec:diff_analysis}.

\subsection{Comparing Different Reference Models}
\label{sec:compar_ref_models}
Our throttle plane analysis is general and can be applied to various reference models. 
Different models encode features 
differently such that they may have
a different level of effectiveness.
We experiment to compare the effectiveness of using VGG16 and Resnet152-Adv as the reference model. 
Note that VGG16 has long been used as a typical feature extractor \cite{neural_style_transfer}, whereas Resnet152-Adv was recently reported as being useful in extracting features \cite{ilyas2019adversarial}. We conduct the same untargeted attack on Resnet50 using these two as the reference model and compare the visual quality (of generated samples) and the level of attack success. In  \autoref{fig:compare_ref_model}, we observe that the throttle planes on the two models have different characteristics.
The throttle plane in VGG16 promotes more natural-looking adversarial samples while being relatively less expressive. In contrast, the throttle plane in Resnet152-Adv allows more harmful perturbation.

\begin{figure*}[ht]
    \centering
    \begin{subfigure}{.49\textwidth}
        \centering
        \includegraphics[width=.9\linewidth]{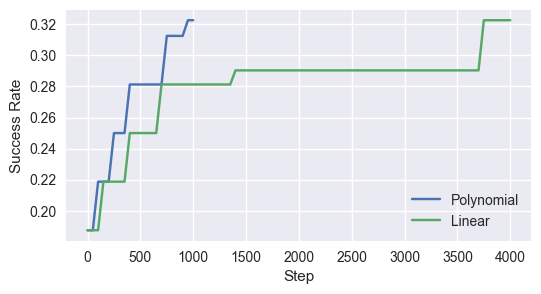}
        \caption{Optimization Speed}
    \end{subfigure}
    \begin{subfigure}{.49\textwidth}
        \centering
        \includegraphics[width=.9\linewidth]{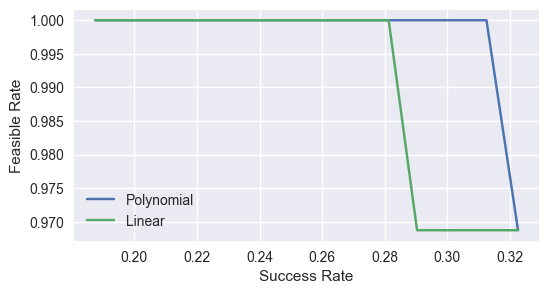}
        \caption{Optimization Feasibility}
    \end{subfigure}
    \caption{Comparison of two barrier loss functions. Graph (a) represents the success rate ($y$-axis) after a given number of steps of optimization ($x$-axis). Graph(b) represents for a given success rate ($x$), how many examples have internal values within their boundary (axis).}
    \label{fig:my_linear}
\end{figure*}

\subsection{Comparing Different Deep Bound}
\label{sec:compare_diff_bound}
In this experiment, we compare the proposed deep distribution bound with the min-max bound mentioned at the beginning of Section~\ref{sec:design}, which is an extension of the fixed pixel range used in BIM and PGD. Different from the motivation example in Section~\ref{sec:design}, here we use our proposed polynomial barrier loss to enforce both bounds instead of naive clipping.
Specifically,
while the  minimum and the maximum of RGB value is $0$ and $255$, we denote the supremum and the infimum activation value at neuron $i$ over a distribution support $\mathcal{S}$ as $\sup_\mathcal{S} y_i$ and $\inf_\mathcal{S} y_i$. The value bound for neuron $i$ is thus as follows. 
\begin{align}
    \ry_i^{\mathrm{adv}} \in \left[\ry_i^{\mathrm{nat}} - \left(\sup_\mathcal{S} \ry_i - \inf_\mathcal{S}{\ry_i} \right) \epsilon,  \ry_i^{\mathrm{nat}} + \left(\sup_\mathcal{S} \ry_i - \inf_\mathcal{S} \ry_i\right) \epsilon \right]
    \label{eq:deepminmaxbound}
\end{align}
It is worth noting that, as the definition implies, the min-max bound only uses the information of the extreme values. While a deep distribution bound uses information from the cumulative distribution function, which is much more informative than just extreme values. Intuitively the rich information encoded in the deep distribution bound gives finer-grained control over the adversarial changes. As in \autoref{fig:SSIM_Human_Bound_Type}, we compare the two bounds under the same setting. Specifically, We attack the Resnet152-Adv using the same VGG16 throttle plane.
All the other settings remain the same except the bound. We measure both the human preference rate and the SSIM score. We observe that the deep distribution bound consistently outperforms the deep min-max bound on the SSIM score. In the human evaluation, for 
the $\epsilon$ relative to BIM4 greater than 20\%, humans consistently prefer the deep distribution bound. And with a smaller scale, the deep distribution bound is slightly worse than the deep min-max bound. We conjecture this could be due to the randomness in human evaluation when the perturbation is at a small scale.

\subsection{Comparing Two Barrier Loss Functions}
\label{sec:two_barrier_loss}
Besides the polynomial barrier loss function, we have also tried a linear barrier loss defined as follows.
\begin{align}
\begin{split}
    \mathcal{L}_i({\ry_i}^{\mathrm{adv}}) = & k\Big\{\relu\left[{\ry_i}^{\mathrm{adv}}-(b\cdot \mathrm{high}+(1-b) \cdot {y_i}^{\mathrm{nat}})\right] +  \relu\left[(b\cdot\mathrm{low}+(1-b)\cdot {y_i}^{\mathrm{nat}}) - {y_i}^{\mathrm{adv}}\right] \Big\}
\end{split}
 \label{eq:linear_barrier_loss}
\end{align} Intuitively, the coefficient $b$ allows us to start applying (linear) penalty when the value approaches the boundaries.
Empirically we set $k=1e6$ and $b=0.95$. It is worth noting that as a common drawback of barrier method, when the $y_i^{\mathrm{adv}}$ goes beyond the feasible boundary, the optimization may encounter numerical issues. However, with a properly setting of $k$ and $b$, most variables can maintain a safe distance from the boundary as we can see in \ref{sec:opt_methods}. Specifically, when $y_i^{\mathrm{adv}}$ is 9\% of the range away from the boundary, only 3\% of samples go out of the boundary. When a numerical exception happens, 
we restore the last feasible sample, decrease the learning rate and resume optimization.

We observe that the linearly growing penalty is not strong enough to discourage bound violation even with a large $k$ value. Figure~\ref{fig:my_linear} presents the polynomial barrier loss both converges faster and constrains the optimization better than the linear loss. 

\subsection{Comparing Optimization Methods}
\label{sec:opt_methods}


As discussed in Section~\ref{sec:poly_barrier_loss}, there are other optimization methods that can be used for our adversarial example generation such as two-step optimization and clipping. We conduct an experiment for those methods in comparison with our polynomial barrier method. The same optimizer, i.e., the Gradient Sign Method, is used for all the methods during evaluation. We use the adversarially trained ResNet152 model as our study subject. The throttle plane used in this evaluation is the plane 
in the last layer of Block 2. We use a bound of 1\% of the difference between the minimum and maximum activation values, which are computed on the entire training dataset. Here, we use a value bound instead of a quantile bound to make different optimization methods comparable.
For the two-step optimization method, we set the number of iterations to 100. At each iteration, we first optimize the internal values once and then the input ten times. We set the step size to 10\% of the internal boundary for the internal optimization, and $2.6e\minus3 \times 255$ for the input optimization.
This optimization setting is similar to that in the paper~\cite{DBLP:conf/ijcai/Kumari0SMKB19}.
For the clipping method, we optimize for 1000 iterations, and at each iteration we update the input once using the step size of $2.6e\minus3 \times 255$. For our algorithm, we run 1000 iterations with the step size of $6.2e\minus3 \times 255$. 


\begin{table}[h]
\centering
\caption{Comparison of optimization methods.}
\begin{tabular}{cccccc}
\toprule
\multirow{2}{*}[-0.3em]{Method}         & \multicolumn{2}{c}{Boundary}     & \multirow{2}{*}[-0.3em]{Consistency} & \multirow{2}{*}[-0.3em]{Succ. Rate} \\ \cmidrule(lr){2-3} 
                                 & Feasibility        & Avg. Size    &                              &                 \\ 
\midrule
\textbf{Polynomial Barrier Method} & \textbf{97\%} & \textbf{91\%} & \textbf{Yes}                 & \textbf{25.8\%}      \\ 
Two-step Optimization            & 0\%               & 204\%            & No                           & 21.9\%               \\ 
Clipping                         & 0\%               & 429\%            & No                           & 21.9\%              \\ 
\bottomrule
\end{tabular}

\label{tab:opt_methods}
\end{table}

\begin{table*}[ht]
\begin{minipage}[t]{0.48\textwidth}
\centering
\scriptsize
\caption{Untargeted attacks on the adversarially trained ResNet152-Adv and detection, with Pref. meaning human preference}
\begin{tabular}{ccccccc}
\toprule
\multirow{2}{*}[-0.3em]{Attack} & \multicolumn{3}{c}{Feature   Squeezing} & \multirow{2}{*}[-0.3em]{JPEG} & \multirow{2}{*}[-0.3em]{\citet{hu2019new}} & \multirow{2}{*}[-0.3em]{Pref.} \\ \cmidrule(lr){2-4}
                                &  2x2    & 11-3-4     & 5-bit         &                       &                                 &                             \\
\midrule
BIM4                           & 50/100     & 55/100       & 57/100      & 57/100                & 48/70                           & 30\%                        \\
FS1                             & 46/100     & 46/100       & 46/100      & 48/100                & 48/70                           & 35\%                        \\
SM50 & 51/100 & 51/100 & 60/100 & 60/100 & 44/70  & 29\% \\
HC 0.1 & 27/100 & 37/100 & 24/100 & 32/100  & \textbf{59/70} & 4\% \\
\system40                & \textbf{79/100}     & \textbf{97/100}      & \textbf{99/100}      & \textbf{64/100}                & 46/70                           & \textbf{41\%}                        \\
\bottomrule
\end{tabular}
\label{tab:untarget_resnet152}
\end{minipage}
\hfill
\begin{minipage}[t]{0.48\textwidth}
\centering
\scriptsize
\caption{Targeted attacks on ResNet50 and detection, with Pref. human preference}
\vspace{0.03in}
\begin{tabular}{ccccccc}
\toprule
\multirow{2}{*}[-0.3em]{Attack} & \multicolumn{3}{c}{Feature   Squeezing} & \multirow{2}{*}[-0.3em]{JPEG} & \multirow{2}{*}[-0.3em]{\citet{hu2019new}} & \multirow{2}{*}[-0.3em]{ Pref.} \\
\cmidrule(lr){2-4}
                                &  2x2     & 11-3-4     & 5-bit         &                       &                                 &                             \\
\midrule                                
BIM4                           & 30/100     & \textbf{91/100}       & 100/100     & 51/100                & 46/70                           & 27\%                        \\
FS1                             & 6/100      & 25/100       & 53/100      & 24/100                & \textbf{65/70}                           & 36\%                        \\
SM50                           & 0/100      & 4/100       & 10/100      & 1/100                & 49/70                           & 21\%                        \\
SM500                           & 5/100      & 72/100       & 95/100      & 46/100                & 48/70                           & 14\%                        \\
Sparse & 0/100 & 0/100 & 2/100 & 0/100 & 27/70 & 20\% \\
HC 0.1 & 0/100 & 1/100 & 1/100 & 1/100 & 57/70 & 15\% \\
\system100               & \textbf{37/100}     & 88/100       & \textbf{100/100}     & \textbf{57/100}                & 60/70                           & \textbf{50\%}                  \\
\bottomrule
\end{tabular}
\label{tab:target_resnet50}
\end{minipage}
\end{table*}

\autoref{tab:opt_methods} illustrates the results. Feasibility denotes the percentage of samples remaining in boundary after the optimization. Average size denotes the average boundary size of all the samples. 
We calculate the size using the equation $\max_{i \in \gI}\left[ \frac{\relu({y_i}^{\mathrm{adv}}-{y_i}^{\mathrm{nat}})}{\mathrm{high} - {y_i}^{\mathrm{nat}}}+ \frac{\relu({y_i}^{\mathrm{nat}}-{y_i}^{\mathrm{adv}})}{{y_i}^{\mathrm{nat}}- \mathrm{low} } \right]$.
Consistency denotes if the target internal  values can be produced in the {\em original model}. Note that these optimization methods insert additional operations (e.g., clipping) that essentially change the dataflow of the original model. An observed internal value in the optimizing model may not be feasible in the original model. Success rate measures the percentage of generated samples that successfully induce misclassifications.
It can be observed that most samples are still feasible after our optimization, while the other two methods cannot enforce the bound. Note that even though the clipping method clips internal values and suppresses gradients, updates on the input can still induce internal values that go beyond bound.
The average boundary size of our method is much smaller than the other two, indicating that our barrier loss function can effectively enforce the bound. Adversarial samples generated by the other two methods are hence much less natural-looking.
For consistency, we observe that the two baseline methods do not have any guarantee.
As the subject model is adversarially trained and the internal bound is very tight, the attack is difficult to succeed.
Nonetheless, 
our method still outperforms the baselines regarding the attack success rate.

\subsection{Evaluation Against Detection Approaches}



We evaluate \system{} against three popular detection approaches: feature squeezing~\cite{xu2017feature}, JPEG~\cite{DBLP:journals/corr/DasSCHCKC17}, and~\citet{hu2019new}. We generate 100 adversarial examples for each of these approaches. For feature squeezing, we use the same settings as in the original paper~\cite{xu2017feature}. For JPEG~\cite{DBLP:journals/corr/DasSCHCKC17}, we compress the image with 75 quality. For \citet{hu2019new}, we use 30 examples for fine-tuning the threshold and the remaining 70 for testing. We compare with three existing attack methods: BIM~\cite{kurakin2016adversarial} (BIM4), feature space attack~\cite{xu2020towards} (FS1) and semantic attack~\cite{semantic_attack} (SM50), whose settings were discussed in Section~\ref{sec:quality}. For all the attacks, we generate untargeted adversarial examples against the adversarially trained ResNet152 and targeted examples against the naturally trained ResNet50, without knowing the existence of detection methods (i.e., not adaptive).
\autoref{tab:untarget_resnet152} and \autoref{tab:target_resnet50} show the results. The three columns for feature squeezing are the results for different defense settings.
We can observe that our untargeted attack \system40 
has the highest human preference.
In the meantime, it achieves better success rates than the other attacks for feature squeezing and JPEG; and  comparable (and high) attack success rates for 
\citet{hu2019new}. Recall that these attacks are on an adversarially trained model and hence the detection techniques may not be able to add much, especially for our attack that closely couples perturbation with existing features.
Similarly for the targeted attacks, with a clearly better human preference rates, 
our attack is more or comparably persistent in the presence of detection.
Note that since this is a normally trained model, some detection techniques such as feature squeezing may provide very good defense. 
Observe that our human preference rates in both scenarios are high, indicating that our attack may potentially conduct more aggressive perturbation to evade detection (e.g., through adaptive attack).
We want to point out while these detection techniques were not designed to guard against our attack, it is still worthwhile to understand how \system{} performs in the presence of these techniques. Detection and defense (e.g., adversarial training) specific for our attack will be the future work.
We have also conducted a transferbility study of our attack. We observe that \systemspace has a comparable/slightly-better transferbility than other attacks. Details can be found in Appendix~\ref{sec:transfer}.


\begin{figure*}[h]
    \centering
        \centering
        \includegraphics[width=1\linewidth]{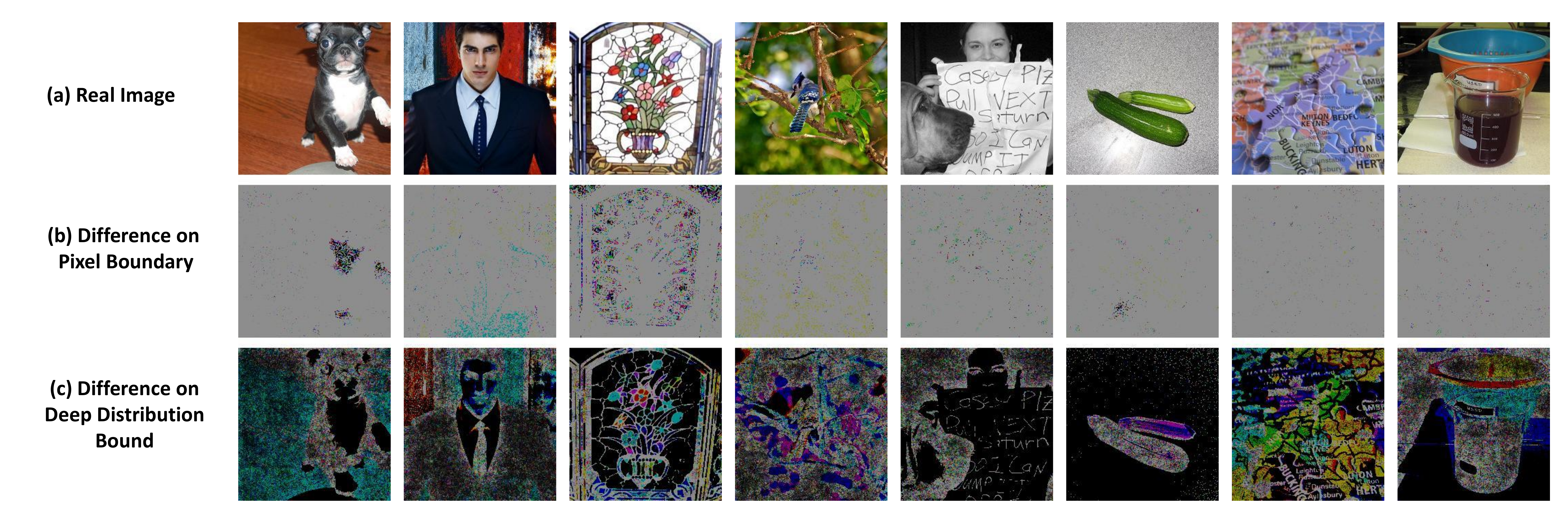}
        \caption{Differential Analysis. The first row denotes benign examples. The following two rows are pixel-wise differences between benign inputs and their corresponding adversarial versions. The second row is from BIM and the third row is from ours. Both have very small bounds. The differences are scaled up by a factor of 2e6 for the two respective rows for better display.}
        \label{fig:differential_analysis}
\end{figure*}

\subsection{Differential Analysis - Understanding the Essence of \system}
\label{sec:diff_analysis}

In this experiment, we give \systemspace
a very small internal bound, i.e., 5e-4\% quantile change on the throttle plane of ResNet50. As such, the pixels changes enabled by the bound indeed are so small that they essentially denote the input gradients induced by our method. We also conduct a similar experiment for the pixel space BIM method (with approximately the same $\ell_\infty=2.77e\minus05 \times 255$) for reference. 
The results are presented in  \autoref{fig:differential_analysis}, where the pixel differences between  adversarial example and their original versions are presented.
We observe that the ``gradients'' in pixel space attacks are more prevalent and uniform, whereas the ``gradients'' in our attack closely couple with the existing content features. 
Note that the experiment cannot be done on feature attack and semantic attack as there is no way to enforce a small bound for those attacks. 

\vspace{-5pt}
\section{Related Work}
\label{sec:related_work}
\vspace{-5pt}
In addition to the works/approaches that we have compared in Section~\ref{sec:experiments}, our technique is also related to the following. Song et al.~\citet{song2018constructing} leveraged a GAN-based method to generate unbounded adversarial examples. Other works propose to uniformly change the colors and lightning conditions for constructing such adversarial examples~\cite{hosseini2018semantic,laidlaw2019functional}. 
Different from those approaches, our \system{} manipulates local content features, which was not studied before.
Xie et al.~\citet{Xie_2019_CVPR} denoised model internal activations, which can achieve better adversarial training results. A study shows that simulating the feature representation of target class images during adversarial generation can increase the transferability of generated samples~\cite{DBLP:conf/cvpr/InkawhichWLC19}.
Some of exiting works utilized the internal representations to facilitate attack~\cite{DBLP:journals/corr/SabourCFF15,DBLP:conf/ijcai/Kumari0SMKB19}. However, they did not consider bounding the activation values in a internal layer.
Specifically, Sabour et al.~\citet{DBLP:journals/corr/SabourCFF15} tried to minimize the $\ell_2$ distance of internal activations between normal and adversarial inputs.
Kumari et al.~\citet{DBLP:conf/ijcai/Kumari0SMKB19} optimized inputs to having a $\ell_\infty$ perturbation on the internal in order to improve adversarial training. In Section~\ref{sec:compare_diff_bound} and \ref{sec:opt_methods}, we have shown that the uniform bound and two-step optimization used in these papers are not effective for our purpose. 
Researchers also tried to perturb the embedding of GAN to generate adversarial samples at which they can directly clip internal activations \citet{disentangle,unrestricted}. However, it is still an open question to obtain a high-fidelity content-preserving bidirectional GAN on ImageNet \cite{bigbigan}.

To improve the imperceptibility of generated adversarial samples, Croce et al.~\citet{sparse} proposed a sparse attack. It defines a salience score for each pixel and avoids large perturbation to salient pixels. HotCold attack~\cite{hotcold} constrains the perturbation of adversarial samples by utilizing SSIM~\cite{SSIM}, a score for measuring structure similarity. However, these scores can be unreliable for bounding perturbations. For example, a study shows that SSIM score can be high (indicting a high similarity) even for unrelated images~\cite{DBLP:journals/corr/abs-1802-09653}. Additionally, while these works focus on imperceptibility, their generated adversarial samples were not evaluated by human studies as we do in~\autoref{fig:humanprefvssucc}. Our attack has stronger attack capability (high attack success rate) while at the same time preserving the imperceptibility (high human preference score).

Some existing defense techniques utilize internal representations. Specifically, DKNN~\cite{dkn} and article~\cite{adv_detect_feat_space} use the distance of internal representations as a criterion to detect adversarial samples.
Defense-GAN~\cite{defense_gan} first learns the manifold of normal samples and then projects adversarial samples to the learned manifold, which is able to remove adversarial noise. Article~\cite{softmax} proposed a feature extractor and used it to detect adversarial samples based on the distance of produced feature representations. However, pointed out by articles~\cite{tramer2020adaptive,DBLP:journals/corr/abs-1802-00420}, it is unclear whether these techniques can effectively defend against adversarial samples whose
perturbations are bounded in deep layers.

\vspace{-5pt}
\section{Conclusion}
\label{sec:conclusion}
\vspace{-5pt}
We propose a novel adversarial attack that can generate natural-looking adversarial examples by bounding model internals. It leverages a per-neuron normal distribution quantile bound and a polynomial barrier loss to handle the non-uniform bounds for internal values. Our evaluation on ImageNet, five models, and comparison with five other state-of-the-art attacks demonstrates that the examples generated by our attack are more natural. It is also more persistent in the presence of various existing detection techniques.

\bibliography{reference}

\newpage
\section{Appendix}
\label{sec:appendix}

\subsection{Transferability of Generated Adversarial Examples}
\label{sec:transfer}

In this section, we study the transferability of adversarial examples generated by different attacks. We launch untargeted attacks on ResNet152-Adv and targeted attacks on ResNet50. We use the same attack success criterion as in Section \ref{sec:quality}: (1) targeted adversarial examples should induce the same target label when transferred to a second model; (2) untargeted adversarial examples should exclude the true label from appearing in the top-5 predicted labels when transferred.
We test generated adversarial samples on 4 different models, including ResNet50' (with the same structure but different parameters as the previously used ResNet50), VGG19, MobileNet and DenseNet. The results can be found in Table \ref{tab:untarget_transfer} and  Table \ref{tab:target_transfer}. We observe that our untargeted attack has higher human preference than other attacks with comparable transferability. For targeted attacks, our method outperforms other methods in both transferability and human preference.
Note that transferring targeted attack is a challenging task and requires specific approaches (e.g., ensemble) to improve the transferability. 

\begin{table}[ht]
\centering
\caption{Transferability of untargeted attacks. Adversarial examples are generated on ResNet152-Adv model.}
\label{tab:untarget_transfer}
\scriptsize
\tabcolsep=2.5pt
\begin{tabular}{ccccccc}
\toprule
\multirow{2}{*}{Attack} & \multicolumn{4}{c}{Transfer to}                & \multirow{2}{*}{Human Preference} \\ \cmidrule(lr){2-5}
                        & ResNet50' & VGG19 & DenseNet121 & MobileNet-V1 &                                   \\ 
\midrule
PGD-4                             & 50/100    & 60/100 & 46/100      & 61/100        & 30\%               \\
FS1                               & 43/100    & 51/100 & 39/100      & 52/100        & 35\%               \\
HC0.1 & 22/100 & 23/100 & 29/100 & 24/100 & 4\% \\
SM50 & 49/100 & 57/100 & 48/100 & 47/100 & 29\% \\
\system40                           & 55/100    & 56/100 & 47/100      & 60/100        & \textbf{41}\%               \\
\system50                           & \textbf{67/100}    & \textbf{63/100} & \textbf{57/100}      & \textbf{70/100}        & 20\%             &  \\
\bottomrule
\end{tabular}
\end{table}

\begin{table}[ht]
\centering
\caption{Transferability of targeted attacks. Adversarial examples are generated on ResNet50 model.}
\label{tab:target_transfer}
\scriptsize
\tabcolsep=2.5pt
\begin{tabular}{cccccc}
\toprule
\multirow{2}{*}{Attack} & \multicolumn{4}{c}{Transfer to}                & \multirow{2}{*}{Human Preference} \\ \cmidrule(lr){2-5}
                        & ResNet50' & VGG19 & DenseNet121 & MobileNet-V1 &                                   \\ 
\midrule
PGD-4                             & 37/100    & 0/100 & 0/100       & 1/100         & 27\%             \\
FS1                               & 17/100    & \textbf{2/100} & 0/100       & \textbf{1/100}         & 36\%             \\
HC0.1 & 1/100 & 0/100 & 0/100 & 0/100 & 15\% \\
Sparse & 0/100 & 0/100 & 0/100 & 0/100 & 20\% \\
SM50                            & 2/100    & 0/100 & 0/100       & 0/100         & 21\%             \\
SM500                             & 32/100    & 0/100 & \textbf{1/100}       & 0/100         & 14\%             \\
\system100                       & \textbf{44}/100    & 0/100 & 0/100       & 0/100         & \textbf{50}\%                 \\
\bottomrule           
\end{tabular}
\end{table}

\subsection{Binary Search for Optimization Step Size}
\label{sec:binarysearch}

In our adversarial example generation, a proper step size (learning rate) is crucial for reliable optimization. A small change on the input can lead to a large quantile change on an internal throttle plane, which makes the optimization osillating and may even lead to numerical exceptions. Choosing an optimal step size depends on model structure and the selected throttle plane(s). It is impossible to manually preset a step size for all the cases. We hence leverage binary search to look for an appropriate step size.

Specifically, we first determine a possible search range for step size, e.g. $[0,255]$ for the gradient sign method on RGB values. We then choose the median value of the search range as a probing step size. We use this probing step size to conduct optimization for a given number of steps.
If the internal quantile change goes beyond the boundary, it means the probing step size is too large for the optimization. We hence update the upper bound of the search range to the current probing value. Otherwise, we update the lower bound with the probing value. We repeat the above search procedure for a given number of iterations. 

\subsection{Choosing Throttle Planes}
\label{sec:choose_throttle_planes}
As we know, different layers represent features of various types, e.g., shallow layers for concrete features and deep layers for abstract features. For imperceptibility, a good idea is to simultaneously harness both concrete features (e.g. local textures) and abstract features (e.g. global outlines). Driven by this intuition, we use multiple throttle planes simultaneously instead of a single one. Guided by our principle of looking for normal distributions, we check the normality of various layers in a model and identify a throttle plane list. We empirically choose three representative throttle planes for each model. We conduct the selection of throttle planes for ResNet152-Adv and VGG16. The specific locations of throttle planes we choose can be found in the following \autoref{tab:locs_of_throttle_model}, and the corresponding distribution samples from these chosen throttle planes can be found at Appendix \ref{appendix: throttledist}. 
Note that although our reference models are these two, the target models can be arbitrary.
For the visual quality studies, we consistently used the throttle planes from VGG16. 

\begin{table}[h]
\centering
\caption{Chosen throttle planes for each model}
\tabcolsep=2pt
\label{tab:locs_of_throttle_model}
\begin{tabular}{cccc}
\toprule
Model         & Plane 1                 & Plane 2                    & Plane 3                    \\
\midrule
ResNet152-Adv & a. the first conv. & b. group 1 & b. group 2 \\
VGG16         & a. conv1\_2          & a. conv2\_2             & a. conv3\_3            \\
\bottomrule
\end{tabular}
\caption*{Notation a. represents right after an operation. Notation b. represents the throttle plane which lies in the last block in the group and before the last ReLU operation; conv. represents a convolution layer.}
\end{table}

\subsection{Evaluation on Different Models and Their Corresponding Distances}
\label{sec:more_models_dist}

In this section, we evaluate \systemspace with different quantile changes on 4 models including DenseNet, MobileNet, VGG19 and ResNet50. We use the same setting as in Section~\ref{sec:quality}. The results are shown in \autoref{tab:target_attack_more_models}. We have similar observations as in \autoref{tab:distance_resnet50} (Section~\ref{sec:quality}). With a similar or higher level of attack confidence, our attack has a smaller $\ell_2$ pixel distance and $\ell_\infty$ quantile distance on all the three planes compared to BIM4. This indicates that our attack is more effective in bounding internal perturbations and can generate more natural-looking adversarial examples.
We also observe that \systemspace induces a larger $\ell_\infty$ pixel distance with a smaller $\ell_2$ pixel distance compared to BIM4 at similar level of attack confidence, which indicates the piggy-backing nature of our attack.

\begin{table}[h]
\centering
\caption{Targeted attacks on various models}
\label{tab:target_attack_more_models}
\scriptsize
\tabcolsep=2.5pt
\begin{tabular}{cccccccc}
\toprule
\multirow{2}{*}[-0.3em]{Models}    & \multirow{2}{*}[-0.3em]{Attack} & \multirow{2}{*}[-0.3em]{Confidence} & \multicolumn{2}{c}{Pixel Distance} & \multicolumn{3}{c}{$\ell_\infty$ Quantile Distance} \\ \cmidrule(lr){4-5} \cmidrule(lr){6-8}
                           &                         &                             & $\ell_2$               & $\ell_\infty$            & Plane 1      & Plane 2      & Plane 3      \\
\midrule
\multirow{6}{*}{MobileNet} & BIM4                    & 47.64                       & 10.51            & 0.04            & 0.58         & 0.78         & 0.88         \\
                           & \system10                  & 26.17                       & 3.22             & 0.05            & 0.06         & 0.08         & 0.09         \\
                           & \system20                  & 39.82                       & 4.86             & 0.06            & 0.11         & 0.15         & 0.17         \\
                           & \system30                  & 44.52                       & 5.55             & 0.07            & 0.17         & 0.23         & 0.26         \\
                           & \system40                  & 47.04                       & 5.92             & 0.08            & 0.22         & 0.30         & 0.34         \\
                           & \system50                  & 47.92                       & 5.99             & 0.08            & 0.28         & 0.38         & 0.43         \\
\midrule
\multirow{6}{*}{DenseNet}  & BIM4                    & 49.21                       & 10.68            & 0.04            & 0.59         & 0.81         & 0.89         \\
                           & \system10                  & 20.60                       & 3.48             & 0.05            & 0.06         & 0.08         & 0.09         \\
                           & \system20                  & 41.14                       & 5.74             & 0.08            & 0.11         & 0.16         & 0.17         \\
                           & \system30                  & 50.94                       & 6.81             & 0.09            & 0.17         & 0.24         & 0.26         \\
                           & \system40                  & 56.33                            &  7.45                &      0.10           &     0.23         &   0.33           &     0.35         \\
                           & \system50                  & 60.10                       & 7.93             & 0.11            & 0.29         & 0.40         & 0.44         \\
\midrule                           
\multirow{6}{*}{VGG19}     & BIM4                    & 56.64                       & 12.13            & 0.04            & 0.68         & 0.88         & 0.97         \\
                           & \system10                  & -2.33                       & 2.82             & 0.04            & 0.07         & 0.09         & 0.09         \\
                           & \system20                  & 7.60                        & 5.58             & 0.08            & 0.13         & 0.17         & 0.19         \\
                           & \system30                  & 18.37                       & 7.97             & 0.11            & 0.20         & 0.26         & 0.29         \\
                           & \system40                  & 29.74                       & 9.87             & 0.12            & 0.27         & 0.35         & 0.38         \\
                           & \system50                  & 40.49                       & 11.59            & 0.14            & 0.33         & 0.44         & 0.48         \\
\midrule                           
\multirow{6}{*}{ResNet50} & BIM4                    & 81.91                       & 10.81            & 0.04            & 0.62         & 0.83         & 0.90         \\
                           & \system10                  & 30.29                       & 4.26             & 0.07            & 0.06         & 0.08         & 0.09         \\
                           & \system20                  & 58.82                       & 6.27             & 0.09            & 0.12         & 0.16         & 0.17         \\
                           & \system30                  & 72.43                       & 7.28             & 0.10            & 0.18         & 0.24         & 0.26         \\
                           & \system40                  & 80.06                       & 7.88             & 0.11            & 0.24         & 0.32         & 0.35         \\
                           & \system50                  & 84.52                       & 8.25             & 0.11            & 0.30         & 0.41         & 0.44        \\
\bottomrule                           
\end{tabular}
\end{table}

\subsection{Adversarial Examples of Different Scales}
\label{appendix: adv_samples_scales}

\begin{figure*}[ht]
    \centering
    \includegraphics[width=.75\linewidth]{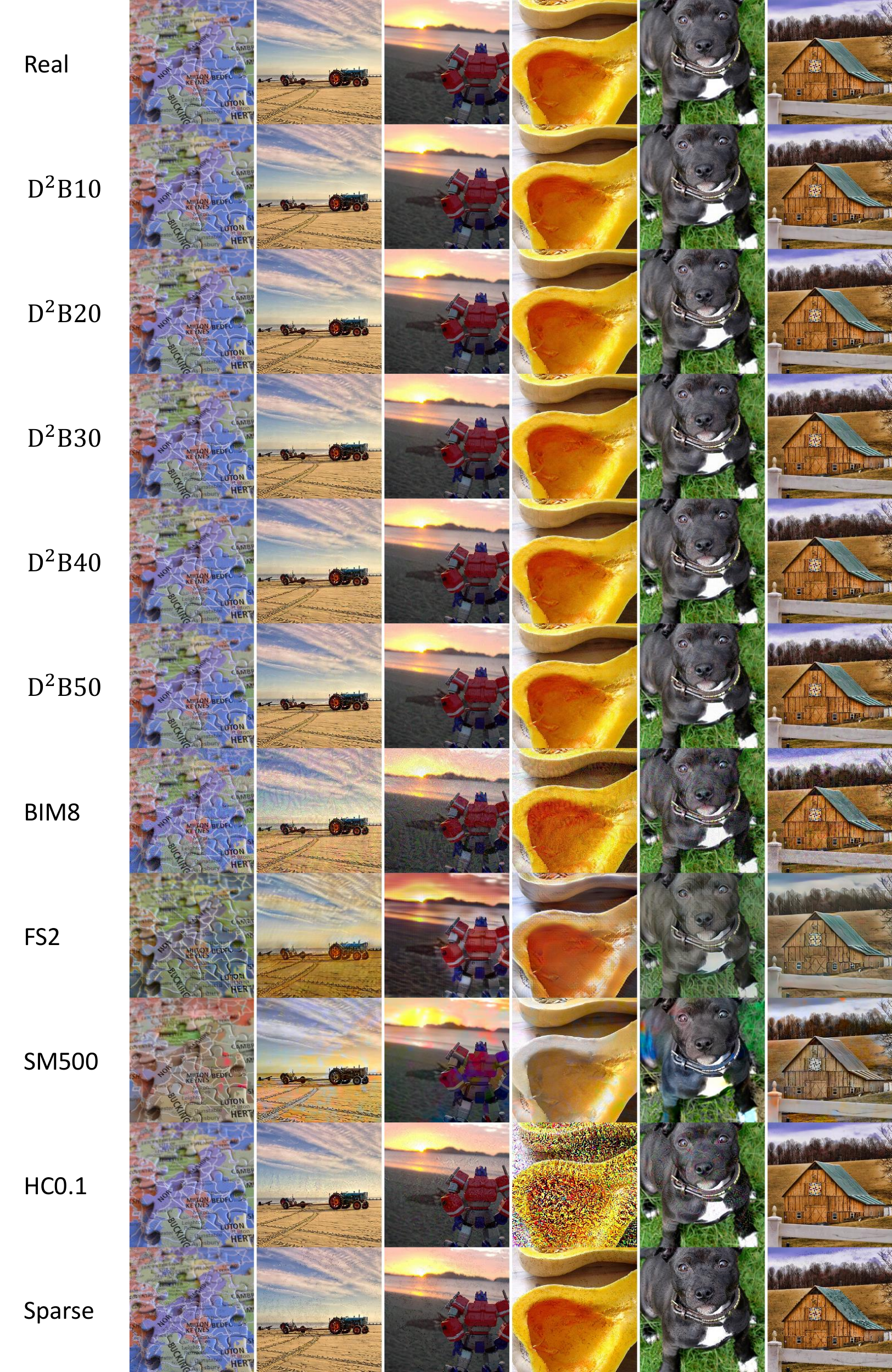}
    \caption{Adversarial samples on ResNet50 from \systemspace of different scales and other Attacks}
    \label{fig:advscales_resnet50}
\end{figure*}

\begin{figure*}[ht]
    \centering
    \includegraphics[width=.75\linewidth]{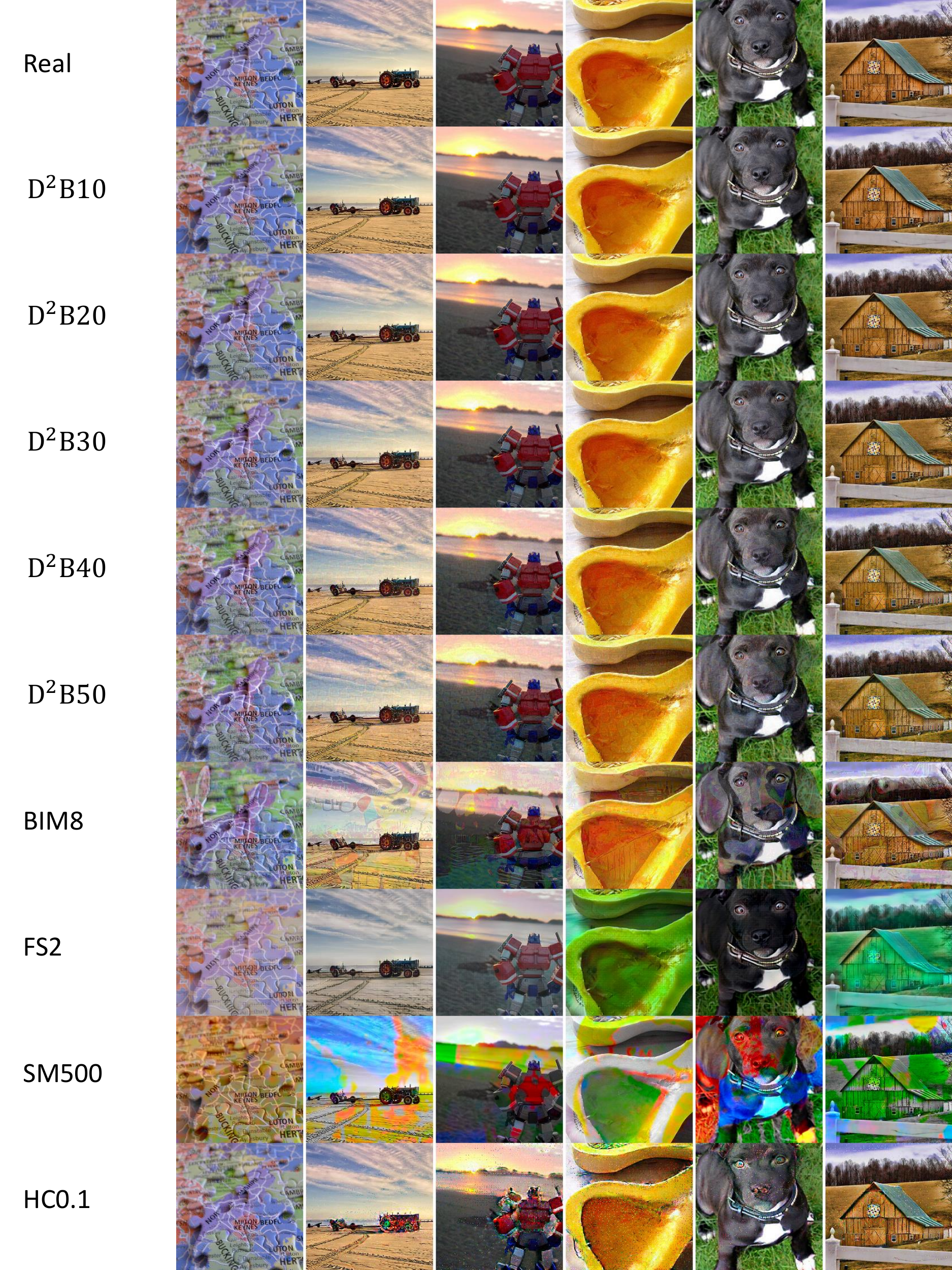}
    \caption{\system Adversarial samples on ResNet152-Adv from \systemspace of different scales and other Attacks}
    \label{fig:advscales_resnet152}
\end{figure*}

We show the generated adversarial examples using \systemspace with different settings in \autoref{fig:advscales_resnet50} and \autoref{fig:advscales_resnet152}.
\autoref{fig:advscales_resnet50} demonstrates samples of a targeted attack on ResNet50 and \autoref{fig:advscales_resnet152} an untargeted attack on ResNet152-Adv. We can observe that most of our adversarial examples are indistinguishable from real images (top row). For few cases such as the 3rd column in the last row (with large quantile change), we observe the presence of a repeating pattern. We speculate this is because the attack was only applied to the first a few representative throttle planes, which
may not be as abstract as other deeper layers.
This effect can be alleviated by including more throttle planes when launching the attack. 


\subsection{Typical Throttle Plane Distributions}
\label{appendix: throttledist}


We present some typical distributions from selected throttle planes in \autoref{fig:typical_dist_vgg16} and \autoref{fig:appendix_renset152_dist}. \autoref{fig:typical_dist_vgg16} shows distribution density graphs of $4\times4\times1$ slice (width$\times$height$\times$channel) of selected throttle planes for VGG16, and \autoref{fig:appendix_renset152_dist} for ResNet152-Adv. 
We observe that the planes from VGG16 resemble normal distributions more than the planes from ResNet152-Adv (e.g., \autoref{fig:dist_vgg16_plane4} and \autoref{fig:dist_resnet152_plane5}). This might explain why adversarial examples from Reference Model VGG16 are relatively more nature.
We also observe that the first few rows and columns in \autoref{fig:dist_vgg16_plane1} and \autoref{fig:dist_resnet152_plane1} look less like a normal distribution. This is due to the existence of zero padding in those layers. The padding operation makes the first a few neurons around the border of a channel distinct from the inner neurons.

\begin{figure*}[ht]
    \centering
    \begin{subfigure}{.6\textwidth}
        \centering
        \includegraphics[width=\linewidth]{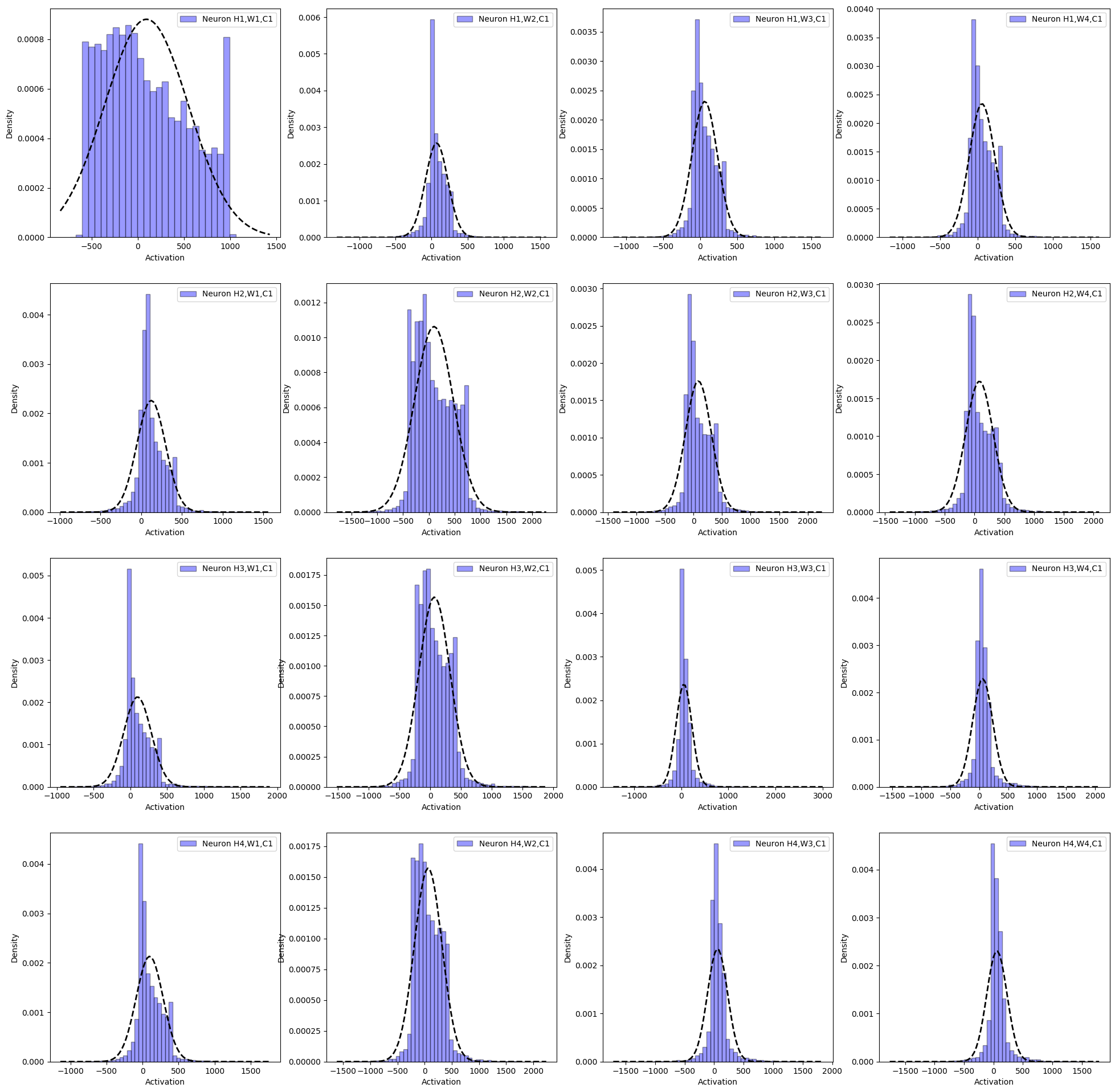}
        \caption{Throttle Plane 1: After VGG16 conv1\_2}
        \label{fig:dist_vgg16_plane1}
    \end{subfigure}
        \begin{subfigure}{.6\textwidth}
        \centering
        \includegraphics[width=\linewidth]{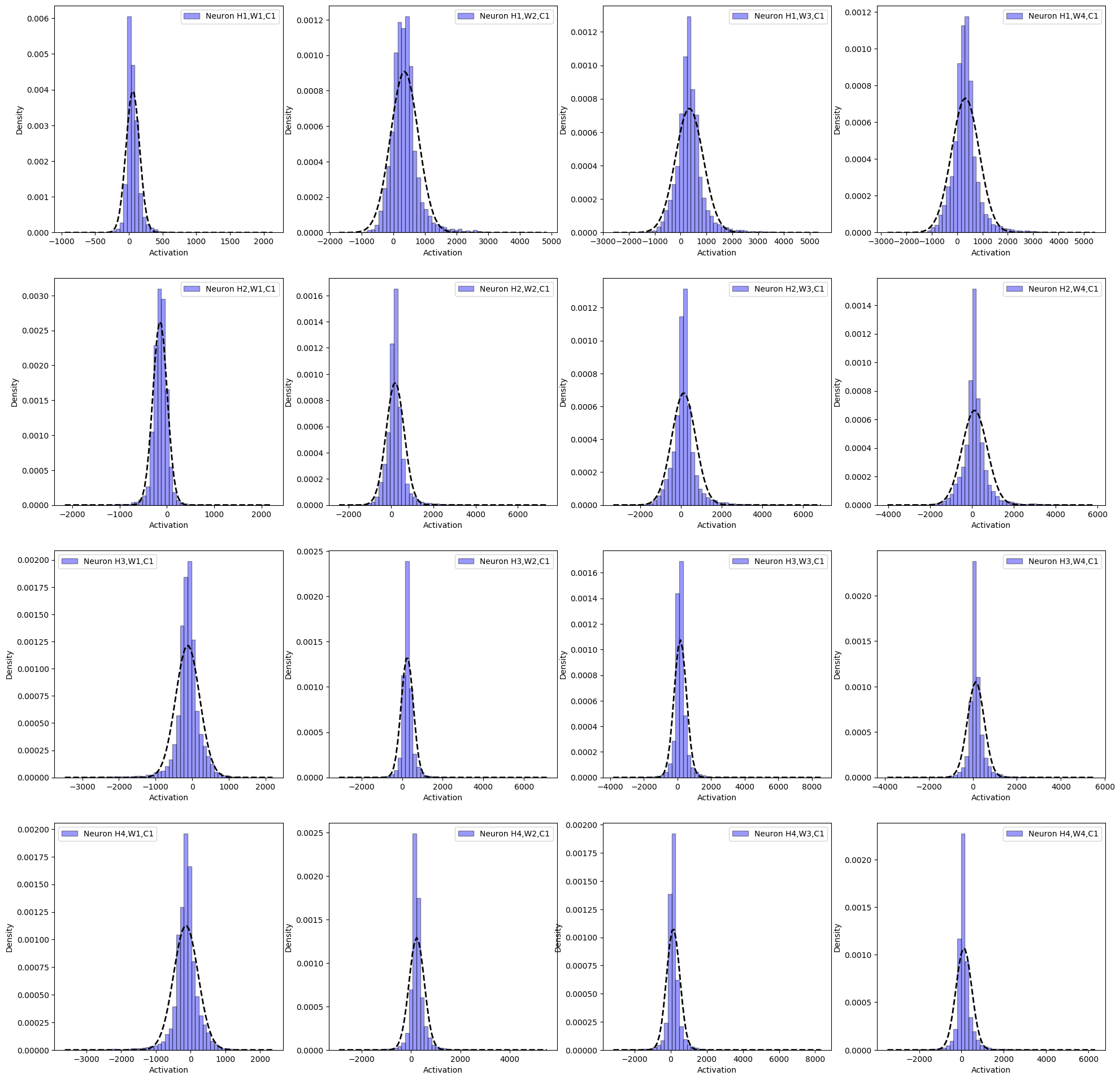}
        \caption{Throttle Plane 2: After VGG16 conv2\_2}
    \end{subfigure}
\end{figure*}
\begin{figure*}[ht]
    \ContinuedFloat
    \centering
        \begin{subfigure}{.6\textwidth}
        \centering
        \includegraphics[width=\linewidth]{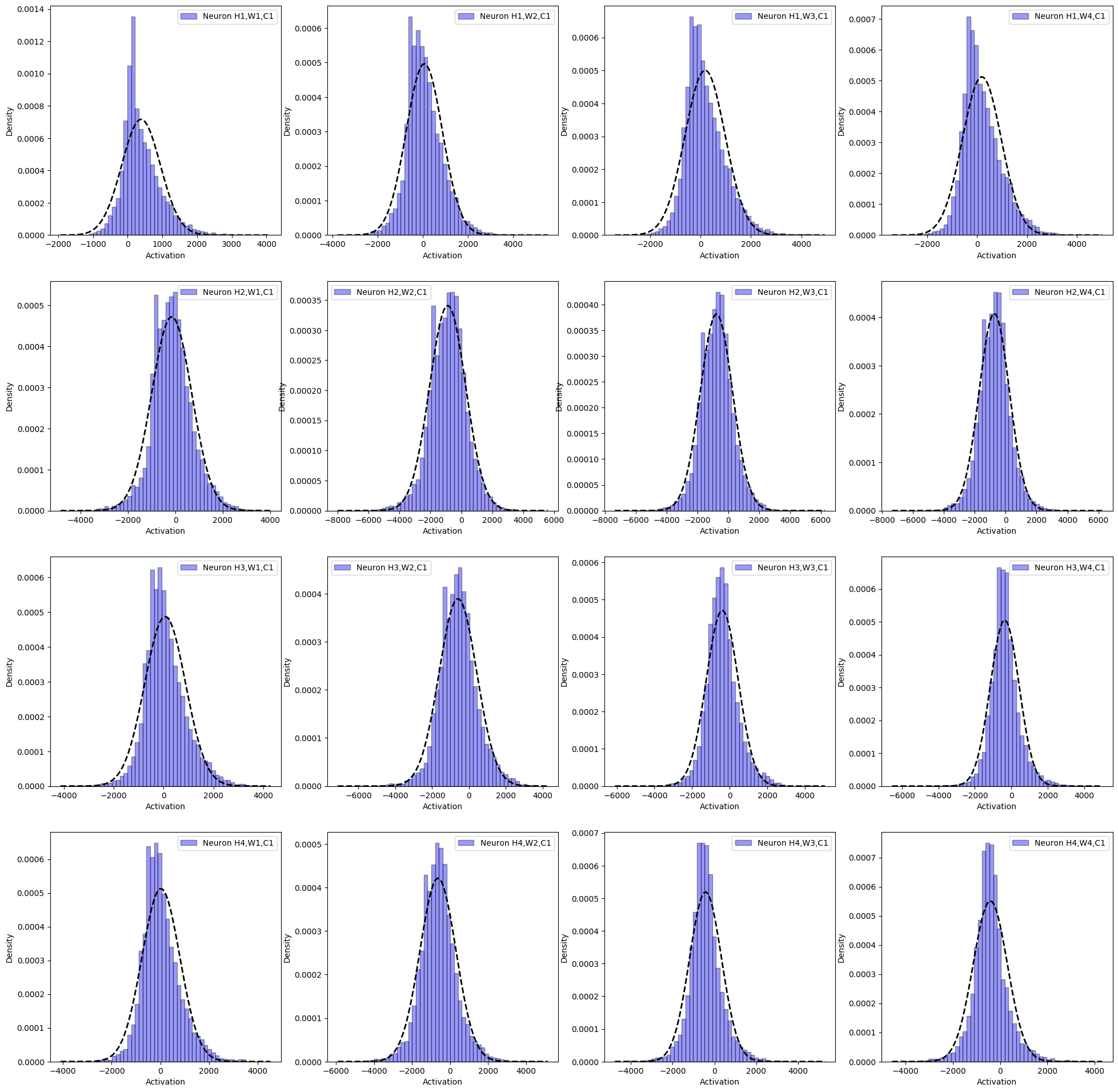}
        \caption{Throttle Plane 3: After VGG16 conv3\_3}
    \end{subfigure}
    \begin{subfigure}{.6\textwidth}
        \centering
        \includegraphics[width=\linewidth]{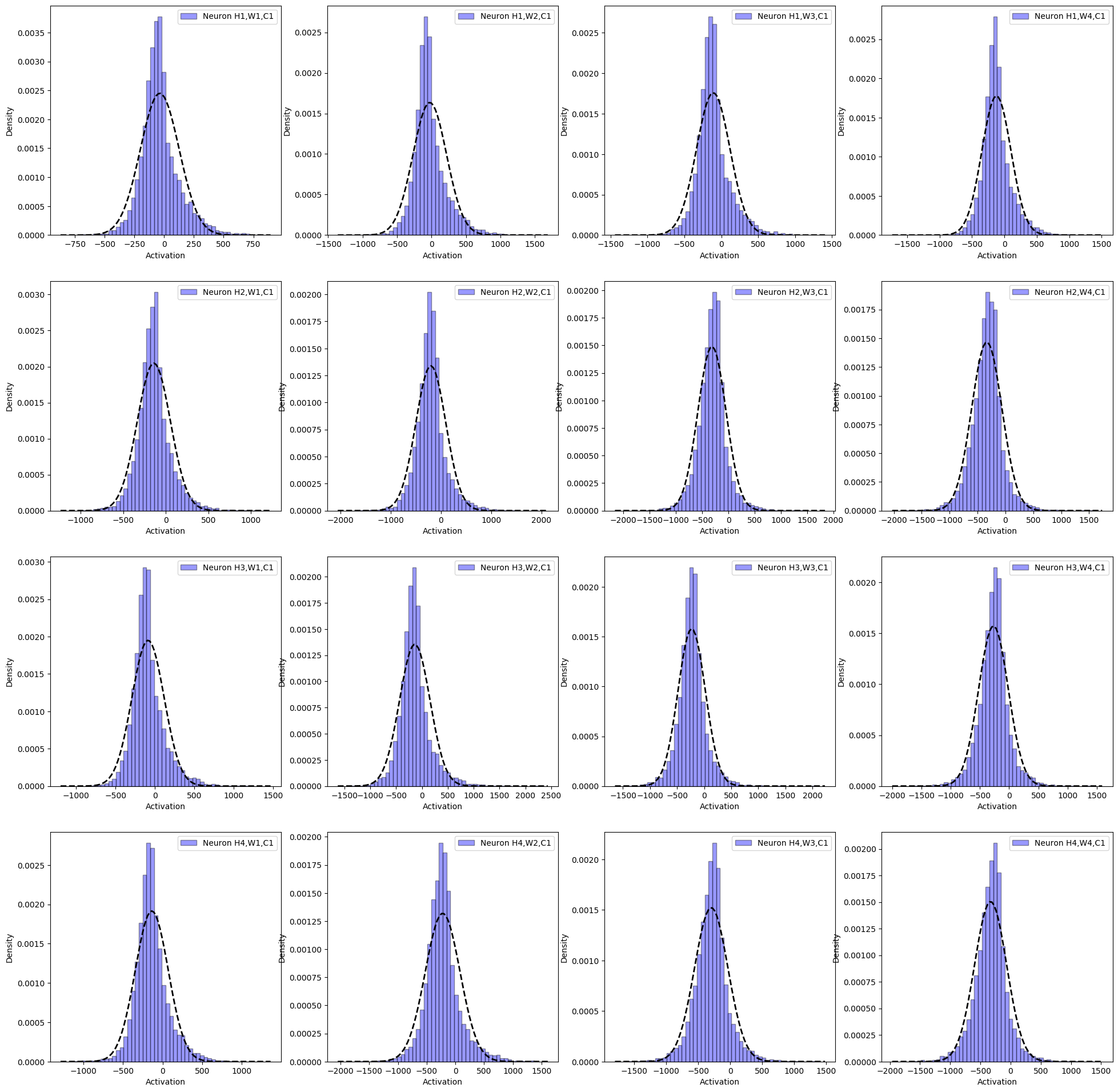}
        \caption{Throttle Plane 4: After VGG16 conv4\_3}
        \label{fig:dist_vgg16_plane4}
    \end{subfigure}
    \caption{Typical Distributions of Selected Throttle Planes from VGG16}
    \label{fig:typical_dist_vgg16}
\end{figure*}

\begin{figure*}[ht]
    \centering
    \begin{subfigure}{.6\textwidth}
        \centering
        \includegraphics[width=\linewidth]{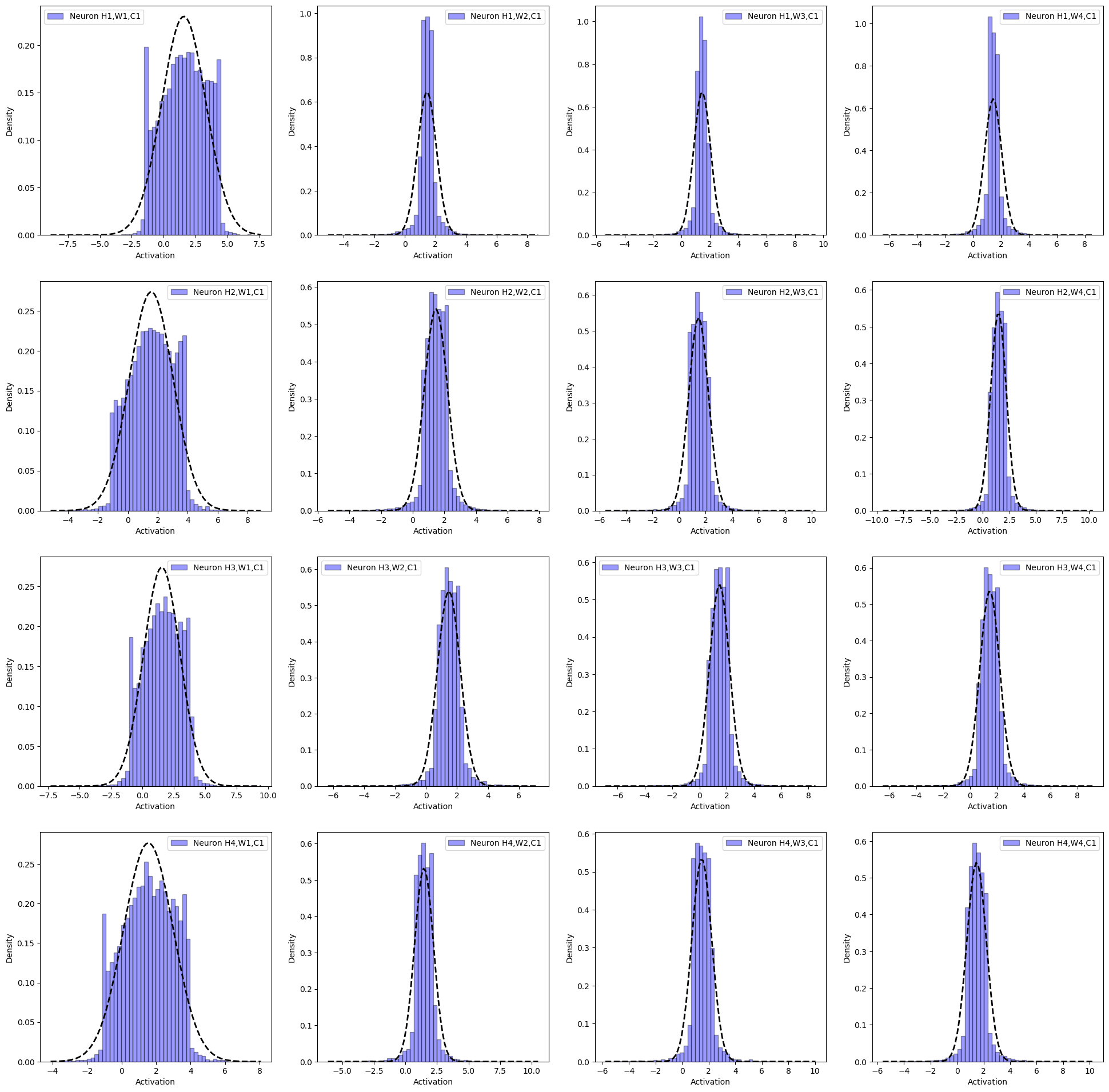}
        \vspace{-10pt}
        \caption{Throttle Plane 1: After the First Convolution}
        \label{fig:dist_resnet152_plane1}
    \end{subfigure}    
    \begin{subfigure}{.6\textwidth}
        \centering
        \includegraphics[width=\linewidth]{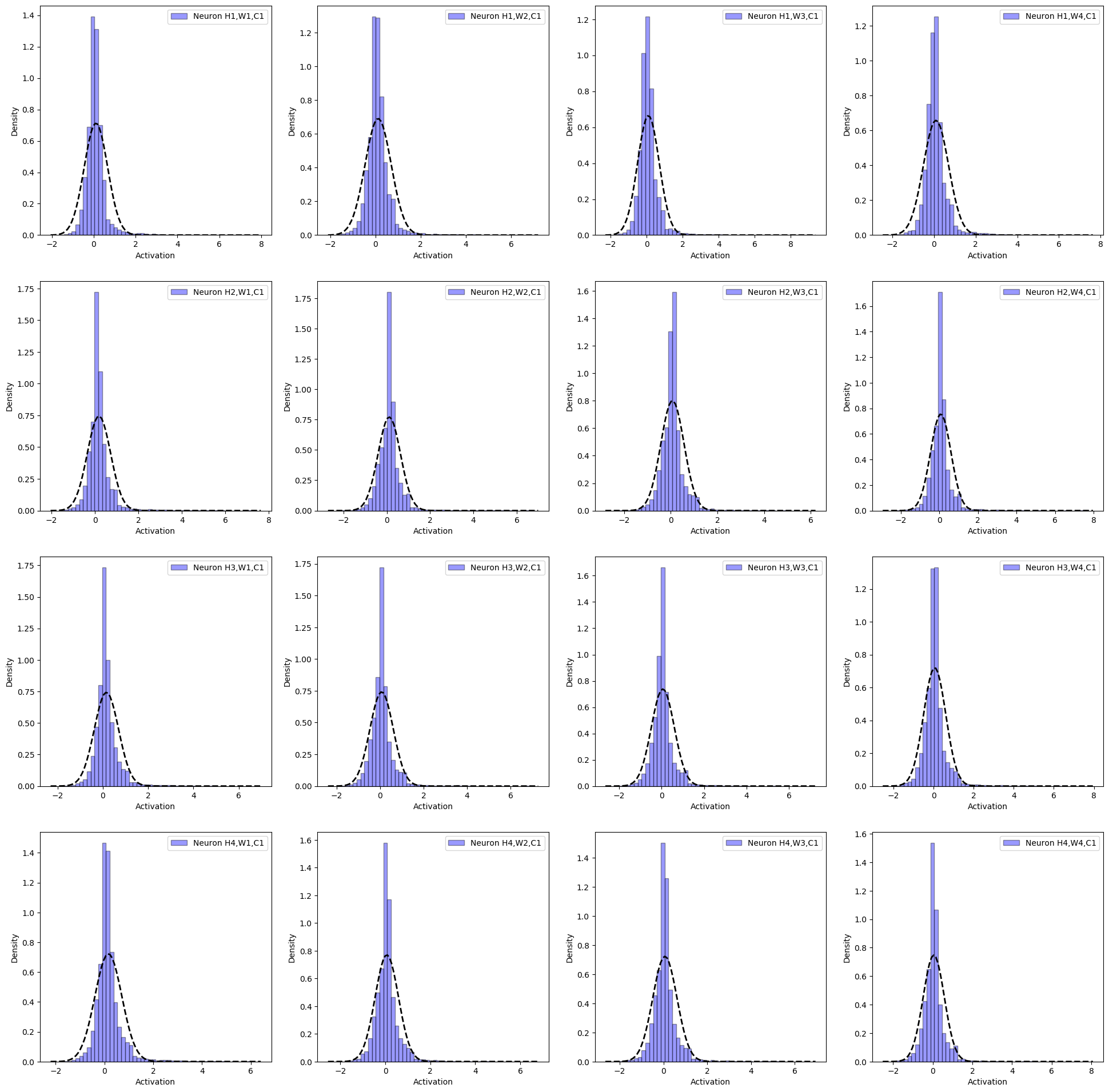}
        \vspace{-10pt}
        \caption{Throttle Plane 2: The last Layer of Group 1 Before ReLU}
    \end{subfigure}
\end{figure*}
\begin{figure*}[ht]
    \ContinuedFloat 
    \centering
    \begin{subfigure}{.6\textwidth}
        \centering
        \includegraphics[width=\linewidth]{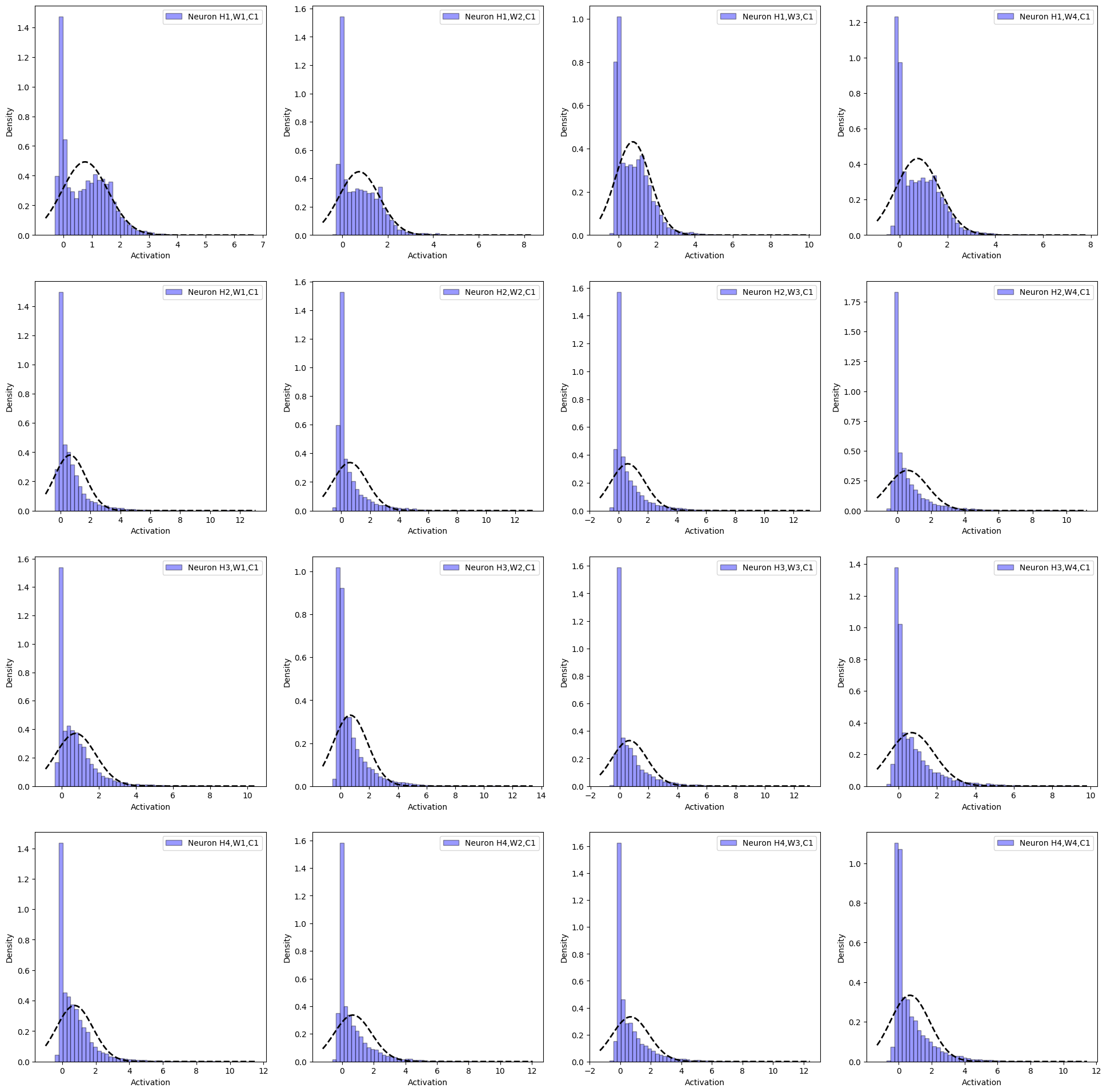}
        \vspace{-10pt}
        \caption{Throttle Plane 3: The last Layer of Group 2 Before ReLU}
    \end{subfigure}
    \begin{subfigure}{.6\textwidth}
        \centering
        \includegraphics[width=\linewidth]{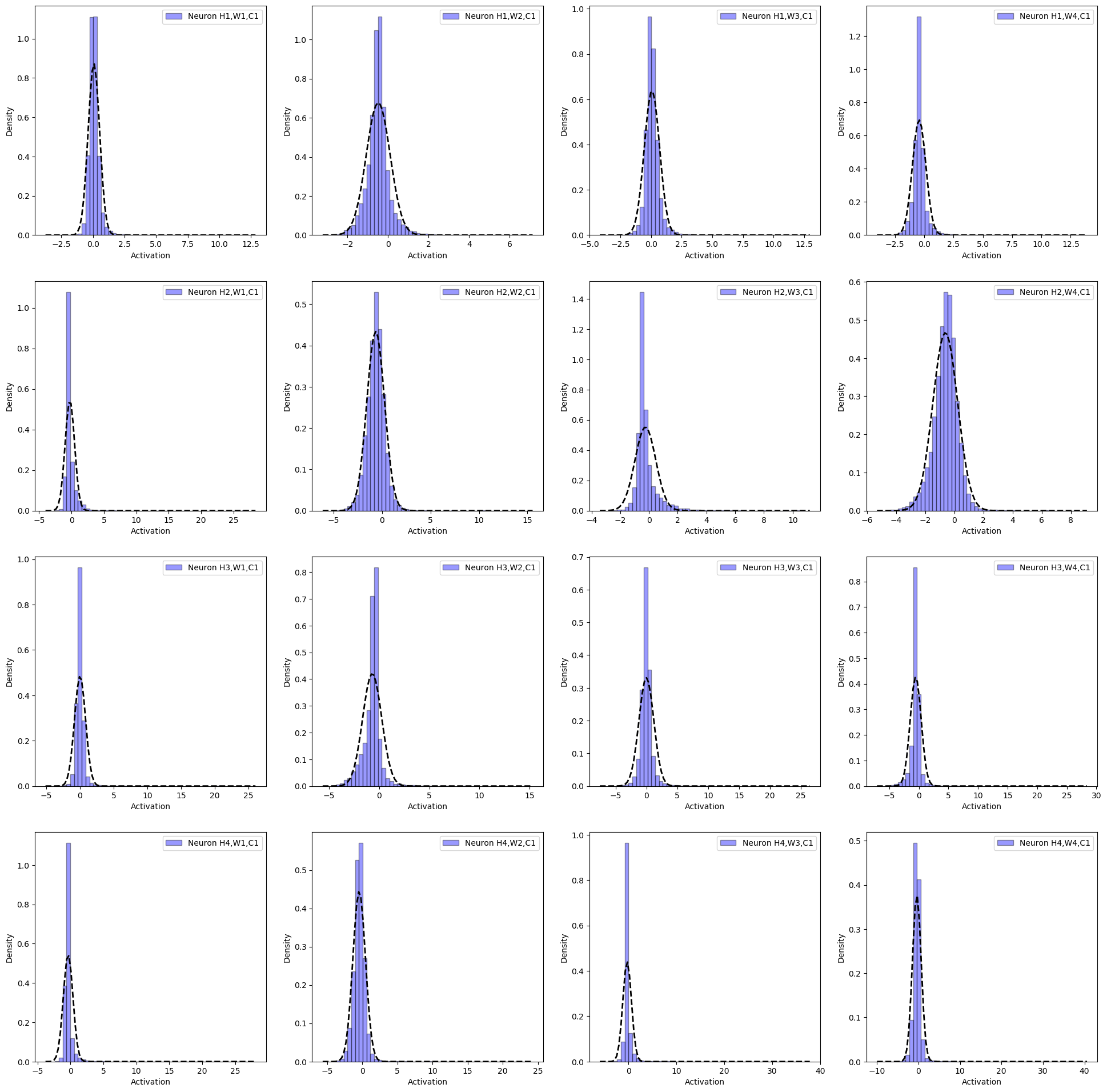}
        \caption{Throttle Plane 4: The last Layer of Group 3 Before ReLU}
        \label{fig:dist_resnet152_plane4}
    \end{subfigure}    
\end{figure*}
\begin{figure*}[ht]
    \ContinuedFloat 
    \centering
    \begin{subfigure}{.6\textwidth}
        \centering
        \includegraphics[width=\linewidth]{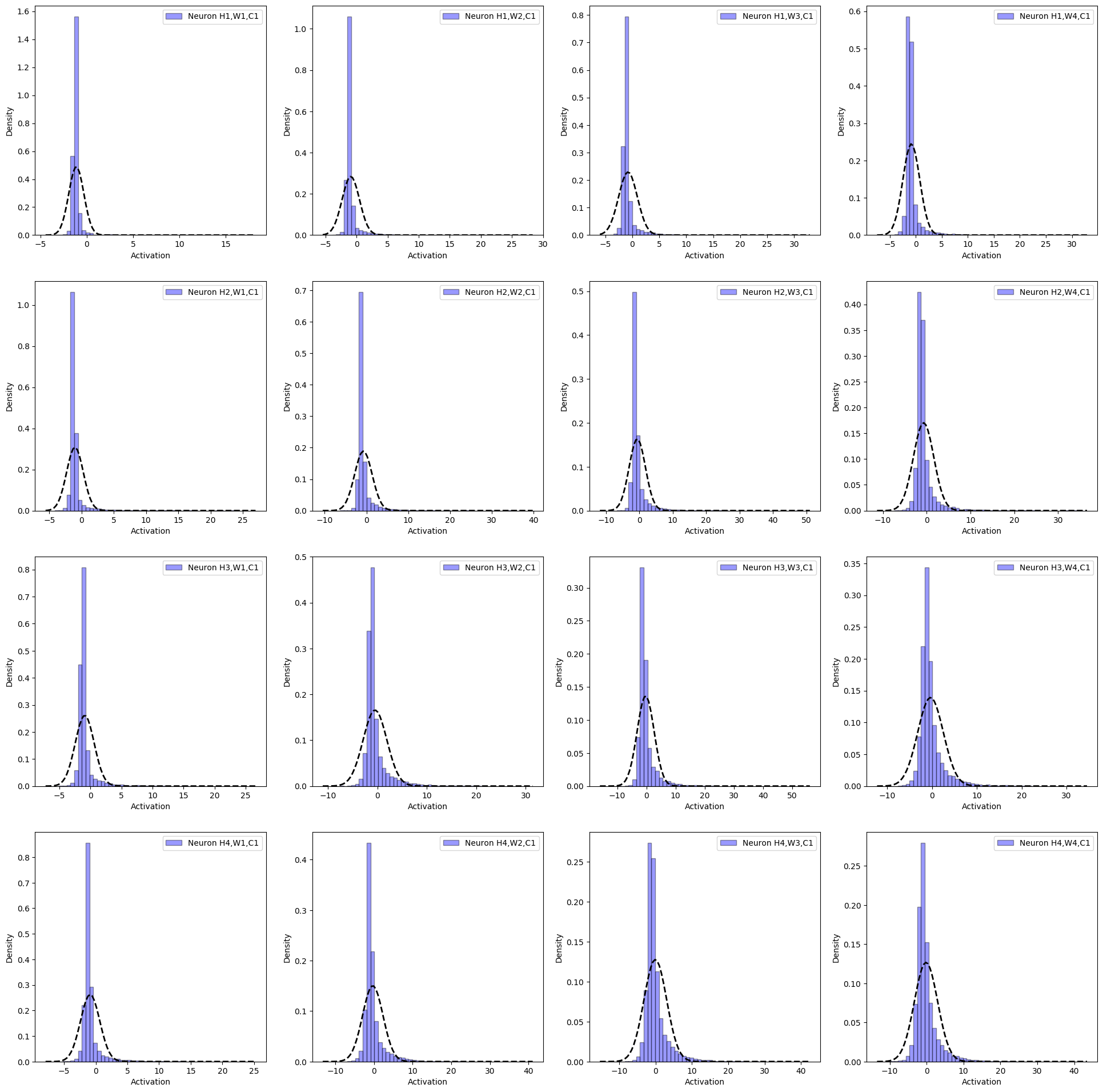}
        \caption{Throttle Plane 5: The last Layer of Group 4 Before ReLU}
        \label{fig:dist_resnet152_plane5}
    \end{subfigure}    
    \caption{Typical Distributions of Selected Throttle Planes from ResNet152-Adv}
    \label{fig:appendix_renset152_dist}
\end{figure*}
\end{document}